\documentclass[review,10pt]{JMtemplate}
\usepackage{bm}
\usepackage{graphicx} 
\usepackage{amsmath,mathrsfs,amssymb} 
\usepackage{amsfonts}  
\usepackage{algorithm}
\usepackage{algorithmic}
\usepackage{siunitx}   
\usepackage{upgreek}   
\usepackage{xcolor}    
\usepackage{subcaption}
\usepackage{booktabs}       
\usepackage{longtable}
\usepackage{threeparttable}
\usepackage{multirow}
\usepackage{multicol}
\usepackage{times}
\usepackage{helvet}
\usepackage{courier}
\usepackage{hyperref}
\usepackage{marvosym}

\usepackage{booktabs}
\usepackage{adjustbox}

\usepackage{natbib}  
\usepackage{caption} 

\begin{document}
\begin{frontmatter}
\title{RMISC: A Large-scale Real-world Multivariate Corpus for Time Series Foundation Models}

\author{\textbf{Qian Sun}\textsuperscript{\rm 1,2,4,*} \qquad \textbf{Yong-Ming Tian}\textsuperscript{\rm 1,2,4,*} \qquad \textbf{Jia-Wei Huang}\textsuperscript{\rm 1,2,4} \\[0.2em] 
\textbf{Cheng Feng}\textsuperscript{\rm 3,4} \qquad
\textbf{Shao-Qun Zhang}\textsuperscript{\rm 1,2,4,~\Letter}
}
\address{
\textsuperscript{\rm 1}State Key Laboratory of Novel Software Technology, Nanjing University, Nanjing, China \\
\textsuperscript{\rm 2}School of Intelligent Science and Technology, Nanjing University, Suzhou, China \\
\textsuperscript{\rm 3} Siemens Data and AI Research, Beijing, China \\
\textsuperscript{\rm 4} Nanjing University – Siemens Joint Research Center on Industrial AI, Suzhou, China
}

\begin{abstract}
Recent years have witnessed the emergence of multivariate modeling using time series foundation models (TSFMs), which achieve advanced zero-shot generalization. Modern multivariate TSFMs are predominantly pretrained on multivariate synthetic data, which is easier to scale but may fail to capture the complex temporal dynamics and cross-variable relationships present in real-world time series. This raises a key question: Whether and to what extent the leading TSFMs trained with the real-world corpus perform better than those trained with synthetic data? To answer this, we establish the RMISC corpus, a considerably large-scale, high-quality, openly accessible, real-world, and multivariate time series archive that contains around 200 datasets and 142 billion time points across diverse domains. Furthermore, we pretrain four advanced TSFMs on univariate, synthetic multivariate, and real-world multivariate data and evaluate their zero-shot generalization capabilities on standard in-distribution and out-of-distribution benchmarks. Experimental results show that incorporating real-world multivariate data predominantly improves the generalization performance for both univariate and multivariate TSFMs. These results provide a deeper understanding of how real-world multivariate data contributes to the development of stronger TSFMs.
\end{abstract}

\begin{keyword}
multivariate time series forecasting \sep time series foundation model \sep real-world time series corpus \sep covariates \sep out-of-distribution generalization
\end{keyword}

\end{frontmatter}

\section{Introduction}\label{sec:Introduction}
Recent advances in Time Series Foundation Models (TSFMs) have significantly remodeled the paradigm of time series analysis~\citep{liang2024foundation}. Fed into large-scale and heterogeneous time series corpora, TSFMs can be directly compatible with diverse forecasting tasks, frequency distributions, and data modalities~\citep{montet2025benchmarking} with remarkable zero-shot generalization capabilities, thus moving beyond traditional statistical methods~\citep{hyndman2018forecasting, box1968some} and deep learning models~\citep{hochreiter1997long, chung2014empirical, sen2019think} that repeatedly train task-specific models for individual time series~\citep{challu2023nhits, lim2021temporal}. In recent years, developers have widely applied TSFM to various fields, such as industrial sensing~\citep{hector2024predictive}, financial assessment~\citep{sezer2020financial}, healthcare monitoring~\citep{morid2023time}, climate modeling~\citep{mudelsee2010climate}, energy management~\citep{hong2016probabilistic}, and traffic prediction~\citep{li2018dcrnn}.

\begin{figure}[t]
    \centering
    \includegraphics[width=1\linewidth]{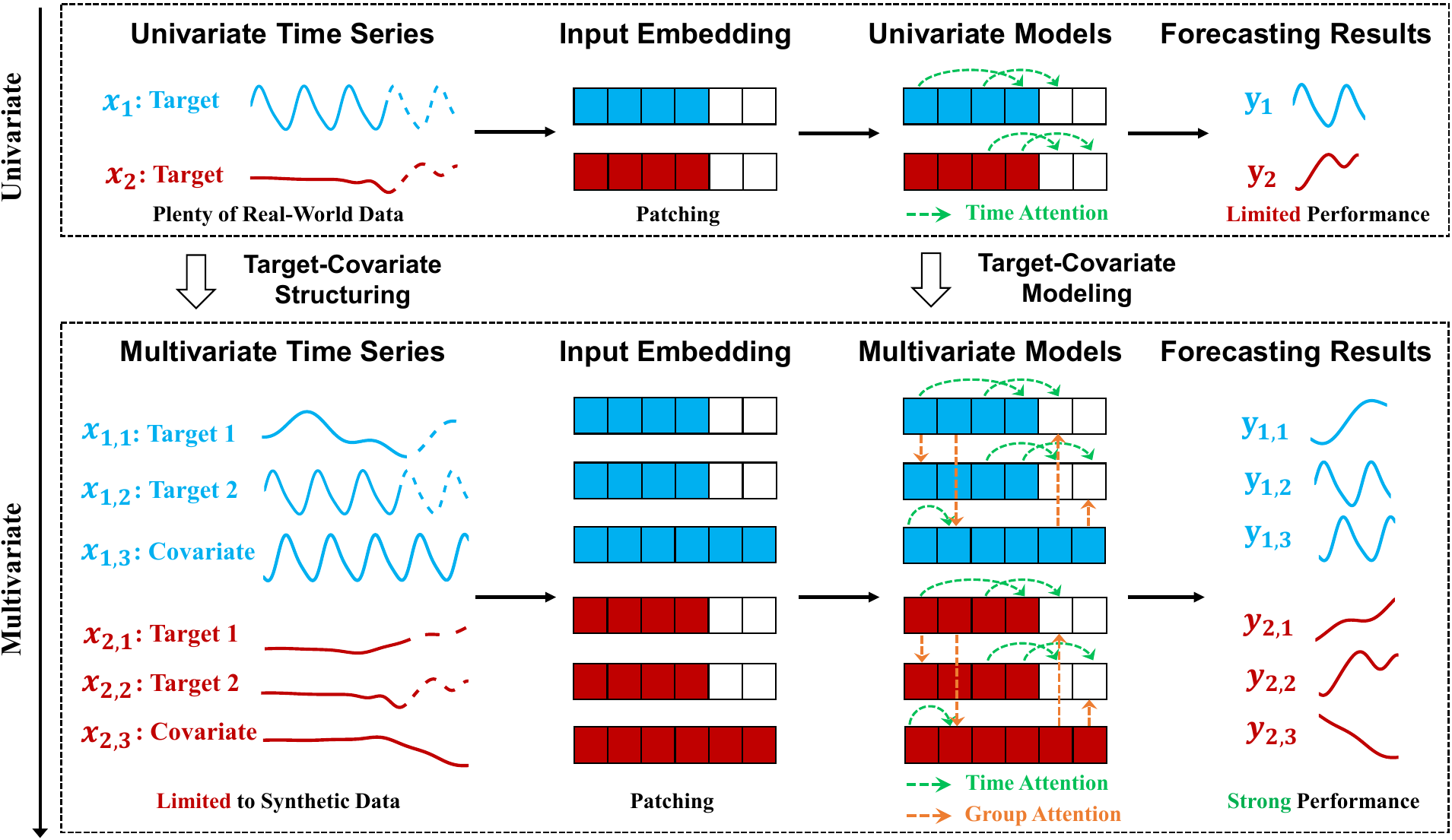}
    \caption{The modeling workflow of univariate and multivariate time series foundation models on corpora.}
    \label{fig:from_univariate_to_multivariate}
\end{figure}

Capturing the cross-variable information is one of the fundamental topics in the development of TSFMs~\citep{ansari2025chronos, liu2025timer}, the modeling workflow of which is illustrated in Figure~\ref{fig:from_univariate_to_multivariate}. Intuitively, real-world time series are rarely observed in isolation; one target variable is usually accompanied by multiple related covariates, and its temporal dynamics are often shaped by complex cross-variable dependencies~\citep{zhang2023crossformer}.  For instance, temperature changes in weather forecasts are affected by rainfall and wind speed. Thus, covariate modeling in multivariate TSFMs contributes to more accurate forecasts as auxiliary covariates and cross-variable dependencies provide complementary signals beyond the target history alone~\citep{lim2021temporal, zhang2023crossformer, liu2024itransformer}. However, current multivariate TSFMs are predominantly pretrained on multivariate synthetic data~\citep{ansari2025chronos, liu2025empowering}; despite the ease of use and scalability, there still exists a gap between synthetic and real-world time series in terms of capturing complex temporal dynamics and cross-variable relationships~\citep{liu2025empowering,li2025uncovering}. This raises a key question: Whether and to what extent do the leading TSFMs trained with the real-world corpus perform better than those trained with synthetic data?

\subsection{Related Studies}
Due to the absence of covariate modeling, the generalization of univariate TSFM remains limited. Recent TSFMs have begun to explicitly incorporate multivariate modeling, involving Chronos-2~\citep{ansari2025chronos}, COSMIC~\citep{auer2025zero}, Toto~\citep{cohen2026time}, GTT~\citep{feng2024only}, TabPFN-TS~\citep{hoo2025tables}, and Moirai-1~\citep{woo2024unified}. Among them, Chronos-2, pretrained on hundreds of millions of multivariate time series data, achieves substantial improvements over univariate TSFMs~\citep{ansari2025chronos, woo2024unified}. Nevertheless, existing real-world multivariate time series datasets still fall short in terms of quantity and quality, challenging the training and evaluation of large-scale multivariate TSFMs~\citep{liu2025empowering}. As an alternative, synthetic time series data has been increasingly explored and used primarily for training multivariate TSFMs~\citep{ansari2025chronos}, which are easier to obtain at scale.

Various synthetic time series generation methods have been explored, ranging from classical statistical models and simulation-based approaches to deep generative models such as GANs, VAEs, and diffusion models~\citep{brophy2023generative, desai2021timevae, yuan2024diffusion}. For example, the Chronos family uses synthetic time series generated by AR and ETS models, TSI, and KernelSynth~\citep{ansari2024chronos, box2015time,hyndman2008forecasting,bahrpeyma2021methodology}. Despite the scalability and flexibility of synthetic time series data, it is often constrained by the assumptions of the generation process and may fail to faithfully preserve real-world complex patterns and complex cross-variable dependencies~\citep{liu2025empowering}. Recent evidence further shows that TSFMs pretrained on synthetic multivariate datasets and performing well on standard benchmarks may still struggle with real-world temporal dynamics~\citep{li2025uncovering}. 

Recent TSFMs have incorporated real-world time series into model development~\citep{woo2024unified, cohen2026time}, and dedicated benchmarks have also been introduced to evaluate models under realistic multivariate forecasting scenarios~\citep{shchur2025fev, aksu2024gift, godahewa2021monash}; however, existing real-world multivariate time series datasets remain limited in quantity and quality, insufficient to fully support the pretraining of large-scale multivariate TSFMs~\citep{pmlr-v235-goswami24a}. Moreover, it is also necessary to build a testbed from multivariate real-world time series data, used to comprehensively evaluate the pretraining and downstream performance of multivariate TSFMs~\citep{shchur2025fev}.

\subsection{Our Contributions}
In this paper, we provide comprehensive investigations on the effects of multivariate TSFMs trained with synthetic and realistic time series data. We establish the Real-world Multivariate tIme Series Corpus (RMISC), which is a considerably large-scale, high-quality, openly accessible, real-world, and multivariate time series archive, as summarized in Table~\ref{tab:rmisc_datasets}. The RMISC corpus contains around 200 datasets and 142 billion time points, collected from real-world scenarios with open and legal licenses, and supports pretraining and benchmarking of multivariate TSFMs.

Furthermore, we empirically compare the convergence and generalization of four advanced TSFMs pretrained on univariate, synthetic multivariate, and real-world multivariate data that corresponds to our proposed RMISC corpus. Specifically, the conducted TSFMs involve Chronos-2~\citep{ansari2025chronos}, GTT~\citep{feng2024only}, Moirai-2.0~\citep{woo2024unified}, and TimesFM-2.5~\citep{das2024decoder}, where the former two are multivariate TSFMs while the latter two are univariate ones. In-distribution performance is measured on in-distribution evaluation sets, while the zero-shot generalization capability is measured on standard out-of-distribution benchmarks that consist of GIFT-Eval~\citep{aksu2024gift} and fev-bench~\citep{shchur2025fev}. As a result, adding real-world multivariate data consistently leads to stronger and more robust performance in out-of-distribution generalization. Specifically, we draw the following conclusions from our experiments: (1) The performance of TSFMs pretrained with multivariate time series consistently outperforms that of univariate data, highlighting the importance of modeling cross-variable dependencies; (2) Replacing synthetic multivariate data with real-world multivariate data yields improvements in both in-distribution and out-of-distribution generalization, potentially benefiting from more realistic temporal dynamics and richer cross-variable dependencies; (3) TSFMs pretrained with a balanced combination of real-world univariate data, synthetic multivariate data, and real-world multivariate data achieve the best overall performance, which we adopt as our final recommended pretraining recipe.

The rest of this paper is organized as follows. Section~\ref{sec:Main_results} introduces the proposed RMISC corpus and its key properties. Section~\ref{sec:Experiments} conducts experiments to investigate how real-world multivariate data affects the performance of pretrained TSFMs. Section~\ref{sec:Discussions} concludes this work.

\begin{table*}[t]
\centering
\caption{A compact summary of RMISC datasets, where ``Obs.'' refers to the total count of time points.}
\label{tab:rmisc_compact}
\scriptsize
\setlength{\tabcolsep}{2.2pt}
\renewcommand{\arraystretch}{0.82}
\begin{adjustbox}{max width=\textwidth, max totalheight=0.92\textheight, keepaspectratio}
\begin{tabular}{@{}llr|llr|llr|llr@{}}
\toprule
\textbf{Dataset} & \textbf{Domain} & \textbf{Obs.} &
\textbf{Dataset} & \textbf{Domain} & \textbf{Obs.} &
\textbf{Dataset} & \textbf{Domain} & \textbf{Obs.} &
\textbf{Dataset} & \textbf{Domain} & \textbf{Obs.} \\
\midrule
ACSF1\cite{gisler2013acsd,schafer2017weasel} & Energy & 0.29 M & CMIP6-2005-PartII\cite{xiaoming2025time,Eyring2016CMIP6} & Environment & 1056.50 M & SP500KnownOpen\cite{sidi2020improving} & Finance & 3.01 M & CSTSNonnormalTrain\cite{degen2025csts} & Others & 151.68 M \\
ApplianceEnergy\cite{candanedo2017appliances} & Energy & 0.51 M & CMIP6-2005-PartIII\cite{xiaoming2025time,Eyring2016CMIP6} & Environment & 1056.49 M & StockFactorsCleaned & Finance & 1133.71 M & CSTSNormalTest\cite{degen2025csts} & Others & 151.83 M \\
AustralianElectricityDemand\cite{godahewa2021australian} & Energy & 1.15 M & CMIP6-2010-PartI\cite{xiaoming2025time,Eyring2016CMIP6} & Environment & 1056.50 M & StockMarketData & Finance & 0.69 M & CSTSNormalTrain\cite{degen2025csts} & Others & 151.68 M \\
AzurePublicDatasetV1\cite{cortez2017resource} & Energy & 3060.08 M & CMIP6-2010-PartII\cite{xiaoming2025time,Eyring2016CMIP6} & Environment & 1056.50 M & TourismMonthly\cite{xiaoming2025time} & Finance & 0.10 M & Car\cite{thakoor2005shape} & Others & 0.07 M \\
AzurePublicDatasetV2\cite{cortez2017resource} & Energy & 4968.71 M & CMIP6-2010-PartIII\cite{xiaoming2025time,Eyring2016CMIP6} & Environment & 1056.49 M & TushareETFDaily & Finance & 24.36 M & CinCECGTorso & Others & 2.33 M \\
BDG2-Bear\cite{miller2020building,xiaoming2025time} & Energy & 1.42 M & ERA5HourlySingleLevels\cite{Nguyen2023ClimateLearn} & Environment & 462.92 M & TushareIndexDaily & Finance & 26.40 M & Covid\cite{hasell2020covid_testing,mathieu2021covid_vaccinations} & Others & 0.01 M \\
BDG2-Fox\cite{miller2020building,xiaoming2025time} & Energy & 2.29 M & GasSensorTemperature\cite{gas_sensor_array_temperature_modulation_487} & Environment & 76.86 M & TushareStockDaily & Finance & 155.79 M & CovidDeaths\cite{godahewa2020covid_deaths} & Others & 0.05 M \\
BDG2-Panther\cite{miller2020building,xiaoming2025time} & Energy & 0.89 M & GlobalClimateChange & Environment & 5.63 M & TushareStockDailyMetrics & Finance & 196.43 M & CovidMobility\cite{godahewa2021covid_mobility} & Others & 0.09 M \\
BDG2-Rat\cite{miller2020building,xiaoming2025time} & Energy & 4.60 M & KDDCup2018\cite{godahewa_2020_4656719} & Environment & 0.54 M & TushareStockWeekly & Finance & 32.64 M & Darts & Others & 0.71 M \\
BatteryRUL & Energy & 0.14 M & OikolabWeather\cite{godahewa_2021_5184708} & Environment & 0.80 M & UKEconomy & Finance & 0.40 M & EMG4Gestures\cite{krilova_2018_emg} & Others & 38.14 M \\
BritainCoal & Energy & 7.96 M & PM25FiveCities\cite{chen2016pm25} & Environment & 1.15 M & WeeklyFuelPricesItaly & Finance & 0.02 M & EbayServer\cite{abdulaal_2021_rt_sync,abdulaal_2021_async} & Others & 3.44 M \\
BuildingsBenchComAmy\cite{Emami2023BuildingsBench} & Energy & 3040.60 M & Subseasonal\cite{mouatadid2023subseasonal} & Environment & 5668.67 M & WeeklyRoadFuelPrices & Finance & 0.002 M & EigenWorms\cite{yemini_2013_celegans} & Others & 27.95 M \\
BuildingsBenchComTmy\cite{Emami2023BuildingsBench} & Energy & 3026.98 M & TemperatureRain\cite{godahewa_2021_temperature_rain} & Environment & 1.17 M & BTS\cite{prabowo2024bts} & Industry & 95.87 M & FordA & Others & 2.46 M \\
BuildingsBenchRealCSV\cite{Emami2023BuildingsBench} & Energy & 39.64 M & Tigge\cite{Rasp_2020} & Environment & 21.01 M & Behavior-1k\cite{li2024behavior} & Industry & 37682.52 M & Gait\cite{multivariate_gait_data_760} & Others & 1.27 M \\
BuildingsBenchResAmy\cite{Emami2023BuildingsBench} & Energy & 4815.70 M & USAirPollution & Environment & 24.45 M & FrothFlotation & Industry & 0.04 M & HAR70Plus\cite{har70+_780} & Others & 15.82 M \\
BuildingsBenchResTmy\cite{Emami2023BuildingsBench} & Energy & 4815.72 M & Weather\cite{godahewa2020weather} & Environment & 14.72 M & GasPipeline\cite{beaver2013scada} & Industry & 1.38 M & HARTH\cite{harth_779} & Others & 27.75 M \\
Bull\cite{xiaoming2025time} & Energy & 0.50 M & WeatherBench5-625deg\cite{Rasp_2020} & Environment & 43783.91 M & GasSensorDynamic\cite{gas_sensor_array_under_dynamic_gas_mixtures_322} & Industry & 37.75 M & HetergeneousHAR\cite{heterogeneity_activity_recognition_344} & Others & 98.90 M \\
Computers & Energy & 0.36 M & WeatherTest & Environment & 1.11 M & LBNL\cite{hong2022three_year_building} & Industry & 122.27 M & HungarianChickenpoxCases\cite{hungarian_chickenpox_cases_580} & Others & 0.01 M \\
ERCOT & Energy & 1.39 M & XiamenAirQuality & Environment & 9.10 M & OccupancyDetection\cite{occupancy_detection__357} & Industry & 0.12 M & Illness & Others & 0.01 M \\
ETT\cite{haoyietal-informer-2021} & Energy & 1.22 M & AMarketChina\cite{wu2022price} & Finance & 3.71 M & PUMP & Industry & 9.69 M & IndoorLocalisation\cite{barsocchi2016multisource} & Others & 1.88 M \\
ETTMulti\cite{haoyietal-informer-2021} & Energy & 1.22 M & AMarketChinaKnownOpen\cite{wu2022price} & Finance & 3.71 M & ProEnFo\cite{wang2023benchmarks} & Industry & 5.31 M & InlineSkate\cite{morchen2006time} & Others & 1.22 M \\
Electricity\cite{trindade_2015_electricity} & Energy & 8.44 M & AliCar & Finance & 0.01 M & RoomOccupancy\cite{room_occupancy_estimation_864} & Industry & 0.17 M & KeplerLightCurves\cite{barbara2022classifying} & Others & 5.89 M \\
ElectricityHourly\cite{godahewa_2020_electricity} & Energy & 8.44 M & Bitcoin & Finance & 2.83 M & SWAT\cite{goh_2017_securewater} & Industry & 7.93 M & LargeST & Others & 4439.10 M \\
GFC2012\cite{xiaoming2025time, wang2023benchmarks} & Energy & 0.50 M & Bizitobs\_application\cite{aksu2024gift,palaskar2024automixer} & Finance & 0.02 M & ServerMachineDataset\cite{su2019robust} & Industry & 21.99 M & M3 & Others & 0.23 M \\
Hog\cite{woo2024unified,xiaoming2025time} & Energy & 0.37 M & Bizitobs\_l2c\_H\cite{aksu2024gift,palaskar2024automixer} & Finance & 0.02 M & SmellSensor & Industry & 402.56 M & M4 & Others & 19.65 M \\
HouseholdPower\cite{individual_household_electric_power_consumption_235} & Energy & 14.53 M & CSI500 & Finance & 643.70 M & WADI & Industry & 23.96 M & MelbournePedestrianCounts\cite{godahewa_2020_4656626} & Others & 3.13 M \\
Ideal\cite{woo2024unified,xiaoming2025time} & Energy & 1.25 M & CausalEffects & Finance & 0.11 M & BeijingSubway\cite{zhang2020deep} & Traffic & 2.98 M & MiniApp\cite{li2024functional} & Others & 0.34 M \\
LondonSmartMeters\cite{godahewa_2020_4656072} & Energy & 71.93 M & ChinaMinuteStock & Finance & 6480.79 M & ChengduTaxi\cite{wang2018will} & Traffic & 2.85 M & MotionSense\cite{Malekzadeh:2019:MSD:3302505.3310068} & Others & 7.42 M \\
OPSD & Energy & 22.90 M & Cif2016-12\cite{woo2024unified,xiaoming2025time} & Finance & 0.006 M & LoopSeattleLA\cite{10.1145/3474717.3483923} & Traffic & 15.89 M & MotorTemperature\cite{wilhelm_kirchgassner_2021} & Others & 15.97 M \\
OPSD-Household & Energy & 47.88 M & Cif2016-6\cite{woo2024unified,xiaoming2025time} & Finance & 0.0006 M & Mdense\cite{de2020spatio} & Traffic & 0.02 M & MZVAV\cite{Granderson2020} & Others & 6.83 M \\
OPSD-PV-Wind\cite{Pfenninger2016PV,Staffell2016Wind} & Energy & 48.74 M & Cryptocurrency & Finance & 9.87 M & Metropt3\cite{metropt-3_dataset_791} & Traffic & 15.73 M & NAB\cite{ahmad2017unsupervised} & Others & 0.32 M \\
OPSD-When2Heat\cite{ruhnau2019heatdemand} & Energy & 45.61 M & CryptocurrencyKnownOpen & Finance & 9.87 M & MetroTraffic\cite{metro_interstate_traffic_volume_492} & Traffic & 0.24 M & PAMAP2\cite{reiss2012pamap2} & Others & 111.72 M \\
OilWell\cite{vargas2019realistic} & Energy & 244.53 M & Dominick\cite{godahewa_2020_4654802} & Finance & 0.51 M & PEMS-Bay-METRO-LA\cite{li2018dcrnn} & Traffic & 24.03 M & Rebound & Others & 120.02 M \\
Pvdaq & Energy & 8.21 M & ExchangeRate\cite{lai2018modeling} & Finance & 0.06 M & PEMSCalifornia\cite{10.1145/3474717.3483923} & Traffic & 38.22 M & Satellite\cite{hundman_2018_spacecraft} & Others & 2.91 M \\
ResidentialPower\cite{bergmeir_2023_8219786} & Energy & 525.09 M & FavoritaSales\cite{favorita-grocery-sales-forecasting} & Finance & 448.49 M & QtrafficSpeed\cite{bbliaojqZhangKDD18deep} & Traffic & 528.77 M & SmartMeterAus30m & Others & 1034.22 M \\
ShellHackathon & Energy & 7.91 M & FavoritaTransactions\cite{favorita-grocery-sales-forecasting} & Finance & 0.25 M & Rideshare\cite{godahewa_2021_5122232} & Traffic & 0.38 M & SmartMeterAus60m & Others & 345.93 M \\
Solar10Minutes\cite{godahewa_2020_solar10min} & Energy & 7.20 M & FavoritaTransactionsKnownOil\cite{favorita-grocery-sales-forecasting} & Finance & 0.25 M & SHandHZMetro\cite{10.1145/3474717.3483923} & Traffic & 20.38 M & SmartMeterUK30m & Others & 500.65 M \\
Solar4Seconds\cite{godahewa_2020_solar} & Energy & 7.40 M & FredMD\cite{godahewa_2020_4654833} & Finance & 0.08 M & T-Drive\cite{Yuan2011Driving,Yuan2010Tdrive} & Traffic & 52.99 M & SmartMeterUK60m & Others & 167.62 M \\
SolarEnergy\cite{lai2018modeling} & Energy & 7.20 M & HierachicalSales\cite{mancuso2021hierarchical} & Finance & 0.42 M & Traffic\cite{lai2018modeling} & Traffic & 15.12 M & StarLightCurves\cite{keogh2006lb_keogh} & Others & 9.46 M \\
TetuanPowerConsumption\cite{salam_2018_tetouan_power} & Energy & 0.42 M & KaggleTS & Finance & 0.05 M & TrafficHourly\cite{godahewa2020_traffic_hourly} & Traffic & 15.12 M & Sunspots & Others & 0.003 M \\
UK-DALE\cite{kelly2015ukdale} & Energy & 65.60 M & M5\cite{m5-forecasting-accuracy} & Finance & 116.21 M & WikiTrafficDaily\cite{godahewa2020kagglewiki} & Traffic & 304.48 M & TimeMMD\cite{liu2024timemmd} & Others & 0.10 M \\
WindElec & Energy & 3.01 M & NIFTYStock & Finance & 4.24 M & WikiTrafficWeekly\cite{godahewa2020kagglewikiweekly} & Traffic & 16.39 M & USBirths\cite{godahewa2020usbirths} & Others & 0.01 M \\
WindFarms\cite{godahewa2020windfarms} & Energy & 19.26 M & NIFTYStockKnownOpen & Finance & 4.24 M & BCI\_Competetion\_IV\_1\cite{blankertz2007noninvasive} & Others & 177.37 M & VehicleTrips\cite{godahewa2021vehicletrips} & Others & 0.0008 M \\
WindPower4secs\cite{godahewa2020windpower} & Energy & 7.40 M & NN5Daily\cite{godahewa_2020_4656110} & Finance & 0.09 M & BCI\_Competetion\_IV\_2a\cite{brunner2008bci} & Others & 143.09 M & WISDM\_V1\cite{kwapisz2010activity,weiss2012personalization} & Others & 3.95 M \\
BeijingAirQuality\cite{chen2017beijing} & Environment & 3.16 M & Restaurant\cite{xiaoming2025time} & Finance & 0.03 M & BCI\_Competetion\_IV\_2b\cite{leeb2008bci} & Others & 25.37 M & WISDM\_V2\cite{kwapisz2010activity,weiss2012personalization} & Others & 10.25 M \\
BeutenbergWeather & Environment & 17.88 M & Rohlik\_orders\_1D\cite{shchur2025fev} & Finance & 0.01 M & BooksPerPerson & Others & 0.01 M & WISDM\_V3\cite{kwapisz2010activity,weiss2012personalization} & Others & 38.88 M \\
CMIP6-2000-PartI\cite{xiaoming2025time,Eyring2016CMIP6} & Environment & 1056.50 M & Rohlik\_orders\_1W\cite{shchur2025fev} & Finance & 0.00 M & BoschCNC\cite{tnani2022smart} & Others & 102.20 M & Worms & Others & 0.23 M \\
CMIP6-2000-PartII\cite{xiaoming2025time,Eyring2016CMIP6} & Environment & 1056.50 M & Rossmann\_1D\cite{shchur2025fev} & Finance & 1.05 M & BrainInvadersBi2014b\cite{korczowski_2019_3267302} & Others & 573.90 M &  &  &  \\
CMIP6-2000-PartIII\cite{xiaoming2025time,Eyring2016CMIP6} & Environment & 1056.49 M & Rossmann\_1W\cite{shchur2025fev} & Finance & 0.15 M & CSE-CIC-IDS2018\cite{Sharafaldin2018ICISSP} & Others & 1266.17 M &  &  &  \\
CMIP6-2005-PartI\cite{xiaoming2025time,Eyring2016CMIP6} & Environment & 1056.50 M & SP500\cite{sidi2020improving} & Finance & 3.01 M & CSTSNonnormalTest\cite{degen2025csts} & Others & 151.83 M &  &  &  \\
\bottomrule
\end{tabular}
\end{adjustbox}
\end{table*}

\section{RMISC Corpus}\label{sec:Main_results}
In this section, we formally introduce the RMISC corpus for both pretraining and benchmarking of multivariate TSFMs. The RMISC corpus is collected from real-world scenarios with open and legal licenses and preserves rich multivariate information with explicit target-covariate annotations. Thus, this corpus can support pretraining and evaluation of multivariate TSFMs under realistic forecasting scenarios where prediction targets, auxiliary covariates, and complex cross-variable dependencies are jointly considered. Table~\ref{tab:rmisc_compact} summarizes the RMISC corpus in terms of dataset name, domain, and total number of observations, and the more detailed information of the RMISC corpus can be accessed from Appendix~\ref{app:full_table}. 

\begin{figure}[t]
    \centering
    \includegraphics[width=1\linewidth]{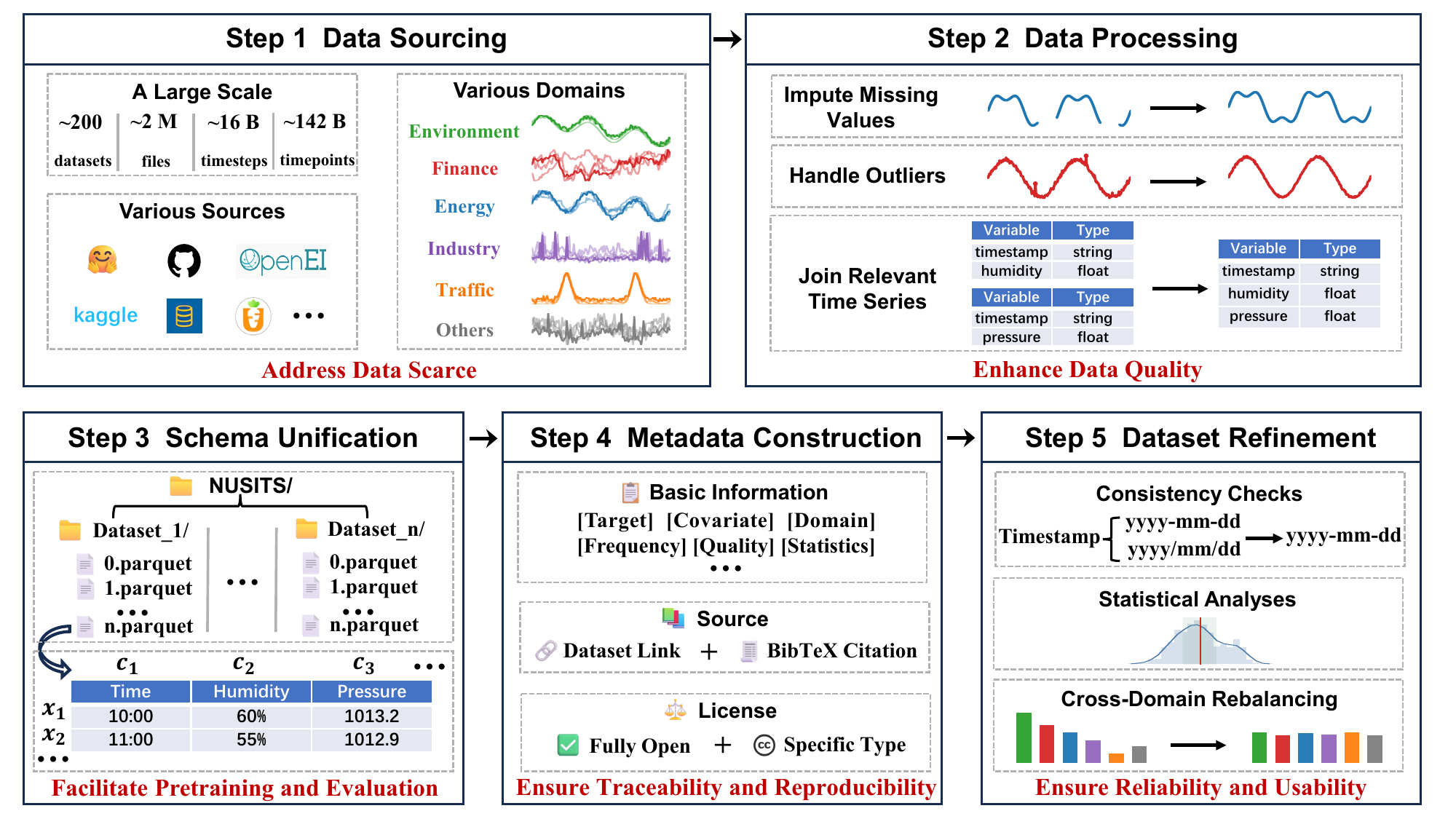}
    \caption{The overall construction pipeline of the RMISC corpus.}
\label{fig:pipeline}
\end{figure}

Constructing the RMISC corpus requires substantial data curation and engineering efforts beyond simple aggregation. Figure~\ref{fig:pipeline} illustrates the overall construction pipeline of the RMISC corpus, involving five key stages, i.e., data sourcing, data processing, schema unification, metadata construction, and dataset refinement.

\noindent\textbf{Stage 1: Data Sourcing.} We first collect a large amount of real-world multivariate time series data from diverse sources and domains. Specifically, the resulting RMISC corpus consists of around 200 sub-datasets, 2 million time-series files, 16 billion timesteps, and 142 billion time points, spanning major real-world domains including energy, finance, environment, industry, traffic, etc.

\noindent\textbf{Stage 2: Data Processing.} Real-world time series data is often noisy and has incomplete information across sources~\citep{hyndman2018forecasting, box2015time}. This step adapts systematic data processing, including handling missing values and outliers, joining correlated time series from multiple files, and transforming raw inputs into consistent time series representations, for enhancing the quality of the collected data. 

\noindent\textbf{Stage 3: Schema Unification.} To facilitate large-scale TSFM pretraining and evaluation, we organize the RMISC corpus in a hierarchical structure, where each subdataset is stored in an independent folder. Within each subdataset, time series data are sequentially partitioned into ordered Parquet files with consistent indexing.

\noindent\textbf{Stage 4: Metadata Construction.} To ensure data traceability and facilitate reproducible research, we design a standardized metadata and provenance system. Each sub-dataset is associated with a metadata file that records prediction targets, covariates, domain, temporal frequency, and other dataset-level statistics. Since RMISC is fully open-source and curated from publicly available real-world multivariate time series datasets, the metadata additionally records the original data source and license information for each sub-dataset. Furthermore, BibTeX citation files are provided whenever formal references are available.

\noindent\textbf{Stage 5: Dataset Refinement.} This step performs overall refinement and validation to further improve the overall reliability and usability of the RMISC corpus. Specifically, we conduct consistency checks across datasets, such as timestamp format standardization. Besides, we perform statistical analyses to assess dataset quality, with detailed results provided in Appendix~\ref{app:statistics}. Note that real-world time series data are inherently unevenly distributed across domains, as privacy-sensitive or commercially valuable sectors such as healthcare and finance often impose stricter constraints on data sharing, licensing, and redistribution~\citep{giuffre2023harnessing}. To address cross-domain imbalance, we construct a balanced version of RMISC by selecting a compact yet domain-balanced subset from the full corpus. The balanced version contains approximately 15 billion time points and follows the same standardized organization as the full dataset.

Together, these five stages ensure that the RMISC corpus is not only a large-scale collection of heterogeneous time series, but also a fully curated, standardized, and benchmark-ready corpus for multivariate TSFM research. Developers can access both the full and balanced versions of RMISC at~\href{https://huggingface.co/datasets/nju-zhangsq/RMISC}{Hugging Face}\footnote{https://huggingface.co/datasets/nju-zhangsq/RMISC}.

\section{Experiments}\label{sec:Experiments}
In this section, we empirically demonstrate the effectiveness of the proposed RMISC corpus. The experiments are performed to answer the question: Whether and to what extent do the leading TSFMs pretrained on the RMISC corpus perform better than those pretrained on univariate and synthetic multivariate data in terms of convergence, in-distribution (ID), and out-of-distribution (OOD) performance?

\subsection{Configurations}\label{subsec:configurations}
\textbf{Datasets.} 
Here, we investigate three types of time series corpora, that is, a real-world univariate corpus, a synthetic multivariate corpus, and our proposed RMISC. The \textbf{R}eal-world \textbf{U}nivariate corpus, denoted as the RU corpus, is derived from the Chronos-2 training corpus. It consists of real-world univariate time series selected from the training corpora of Chronos~\citep{ansari2024chronos} and GIFT-Eval~\citep{aksu2024gift}, comprising approximately 55B univariate time points. The \textbf{S}ynthetic \textbf{M}ultivariate corpus, denoted as the SM corpus, is constructed following the synthetic data construction pipeline of Chronos-2 and comprises approximately 150B time points. Since the exact synthetic multivariate corpus used in Chronos-2 is not publicly released, we reproduce this pipeline to construct our own synthetic multivariate time series dataset. Specifically, we first generate base univariate time series using autoregressive (AR) models, exponential smoothing (ETS) models, TSI, and KernelSynth~\citep{ansari2024chronos,box2015time,hyndman2008forecasting,bahrpeyma2021methodology}. We then apply multivariatizers to these base time series, introducing contemporaneous and sequential dependencies to obtain multivariate time series that form the SM corpus. The proposed RMISC corpus serves as the \textbf{R}eal-world \textbf{M}ultivariate corpus, denoted as the RM corpus.

For each corpus, we randomly sample 20M instances for pretraining using an 80\% rule. Specifically, for subdatasets with more than 10 time-series files, we apply a file-level split, where all time steps from the first 80\% of time-series files are used for training. For the remaining subdatasets, where a file-level split would be less reliable due to the limited number of files, we apply a temporal split, using the first 80\% of time steps in each time series for training. Based on the sampled RU, SM, and RM corpora, we construct seven training corpora corresponding to all non-empty subsets of the three sources, including three single-source corpora, three two-source combinations, and one three-source combination.

\begin{table}[t]
\centering
\caption{Configurations of model architecture and pretraining, where $d_{\text{model}}$, $d_{\text{ff}}$, and $d_{\text{kv}}$ denote the embedding dimension, hidden dimension of feed-forward networks, and key-value dimension, respectively.}
\label{tab:model_pretraining_configs}
\resizebox{\linewidth}{!}{
\begin{tabular}{l|c c c c c c c|c c c}
\hline
\multicolumn{1}{c|}{\textbf{Model}} 
& \multicolumn{7}{c|}{\textbf{Model Architecture Configuration}} 
& \multicolumn{3}{c}{\textbf{Model Pretraining Configuration}} \\
\cline{2-8} \cline{9-11}
\multicolumn{1}{c|}{} 
& \textbf{Modeling Type} 
& \textbf{Layers} 
& \textbf{$d_{\text{model}}$}
& \textbf{Heads} 
& \textbf{$d_{\text{ff}}$}
& \textbf{$d_{\text{kv}}$}
& \textbf{Size} 
& \textbf{Learning Rate} 
& \textbf{Batch Size} 
& \textbf{Optimizer} \\
\hline
Chronos-2 & Multivariate & 12 & 768 & 12 & 3072 & 64 & $\sim$120M & 1e-4 & 64 & AdamW \\
GTT & Multivariate & 8 & 512 & 12 & 3072 & 64 & $\sim$70M & 1e-4 & 32 & AdamW \\
Moirai-2.0 & Univariate & 12 & 768 & 12 & 3072 & 64 & $\sim$120M & 1e-3 & 256 & AdamW \\
TimesFM-2.5 & Univariate & 10 & 1024 & 16 & 1024 & 64 & $\sim$70M & 1e-4 & 768 & AdamW \\
\hline
\end{tabular}
}
\end{table}

\textbf{Models.} We consider four representative TSFMs, including Chronos-2~\citep{ansari2025chronos}, GTT~\citep{feng2024only}, Moirai-2.0~\citep{woo2024unified}, and TimesFM-2.5~\citep{das2024decoder}. Chronos-2 and GTT are multivariate TSFMs trained with multivariate inputs and can explicitly incorporate covariates, whereas Moirai-2.0 and TimesFM-2.5 follow a univariate modeling paradigm. For the multivariate models, target variables and covariates are provided according to their native multivariate input formats. For the univariate models, each multivariate time series is decomposed into multiple univariate series, which are then treated as independent training instances, neglecting the corresponding covariates and cross-variable dependencies. Specifically, for TimesFM-2.5, although pretraining and validation are conducted in a univariate manner, we use its XReg interface during the downstream OOD benchmark to incorporate available covariates, which adjust the model forecasts using external regressors. Table~\ref{tab:model_pretraining_configs} lists the recommended model settings of the four TSFMs.

We conduct separate pretraining runs for each TSFM on the seven training corpora, where the pretraining task is formulated as forecasting future values from historical observations. To focus the comparison on the effect of different training corpora, we keep the overall pretraining protocol consistent with Chronos-2. For each training instance, we randomly crop a context window from the original time series, with the context length ranging from 64 to 1984, and use it to predict the subsequent 64 time steps. This strategy exposes the models to diverse context lengths during pretraining and helps maintain their performance on shorter time series. To ensure consistency of multivariate inputs, we restrict the maximum number of channels for a time series sample to 24, including targets and covariates. To achieve a unified numerical magnitude for time series samples across different datasets, we apply robust instance normalization to each training instance. Specifically, we standardize both the historical context and the prediction window of each variable using the mean and standard deviation computed from the historical context. Then, we apply an inverse hyperbolic sine transformation to reduce the influence of extreme values. Table~\ref{tab:model_pretraining_configs} provides further details of the pretraining settings of the models.

\textbf{Evaluation.} 
Our evaluation includes ID and OOD testing. For ID evaluation, we randomly sample 5M instances from the held-out portion of each corpus as the validation dataset, which corresponds to the remaining 20\% after constructing the training split. Since different TSFMs adopt different training objectives, we measure ID performance using the native loss function of each pretrained model. Specifically, Chronos-2 is evaluated with Sum Quantile Loss (SQL), GTT with Huber Loss (HL), Moirai-2.0 with Weighted Quantile Loss (WQL), and TimesFM-2.5 with a combination of HL and WQL. For OOD evaluation, we evaluate the pretrained TSFMs on two widely used time series forecasting benchmarks, that is, GIFT-Eval~\citep{aksu2024gift} and fev-bench~\citep{shchur2025fev}. To ensure a fair evaluation and avoid potential data leakage, all benchmark datasets overlapping with the pretraining data are excluded, and the remaining are used for OOD evaluation. All models are evaluated directly without dataset-specific fine-tuning; the resulting forecasts reflect the zero-shot OOD generalization capability of the pretrained TSFMs. We also split each benchmark into univariate and multivariate subsets. When the prediction horizons become longer than the native output length of the pretrained models, we employ autoregressive rolling prediction. Following the standard evaluation protocols of these benchmarks, we report mean absolute scaled error (MASE) for point forecasting and WQL for probabilistic forecasting. All experiments are conducted on NVIDIA RTX 5090 $\times$ 8 and 6000 Ada $\times$ 8.

\subsection{In-distribution Forecasting}
\begin{figure}[t]
    \centering
    \includegraphics[width=0.95\linewidth]{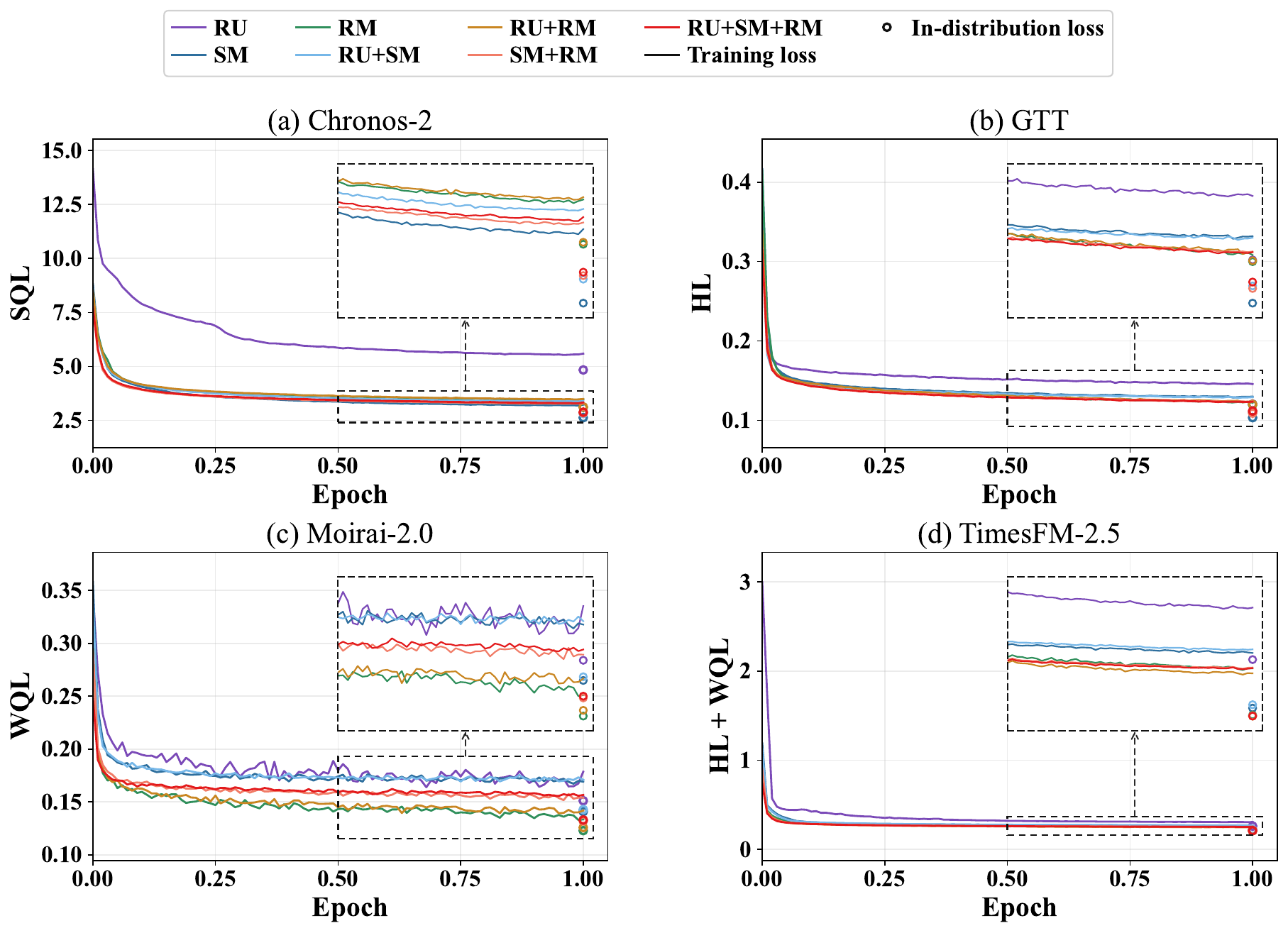}
    \caption{Training and ID loss curves of four TSFMs on different training corpora of the first epoch.}
    \label{fig:first_epoch_train_val}
\end{figure}

To focus the comparison on the effect of different training corpora, we evaluate the models on the held-out ID set of the same corpus used for pretraining. Since ID evaluation preserves the original training objective of each model, we compare ID results within each model across different pretraining corpora and training progress, rather than directly comparing results across models.

\begin{figure}[t]
    \centering
    \includegraphics[width=0.95\linewidth]{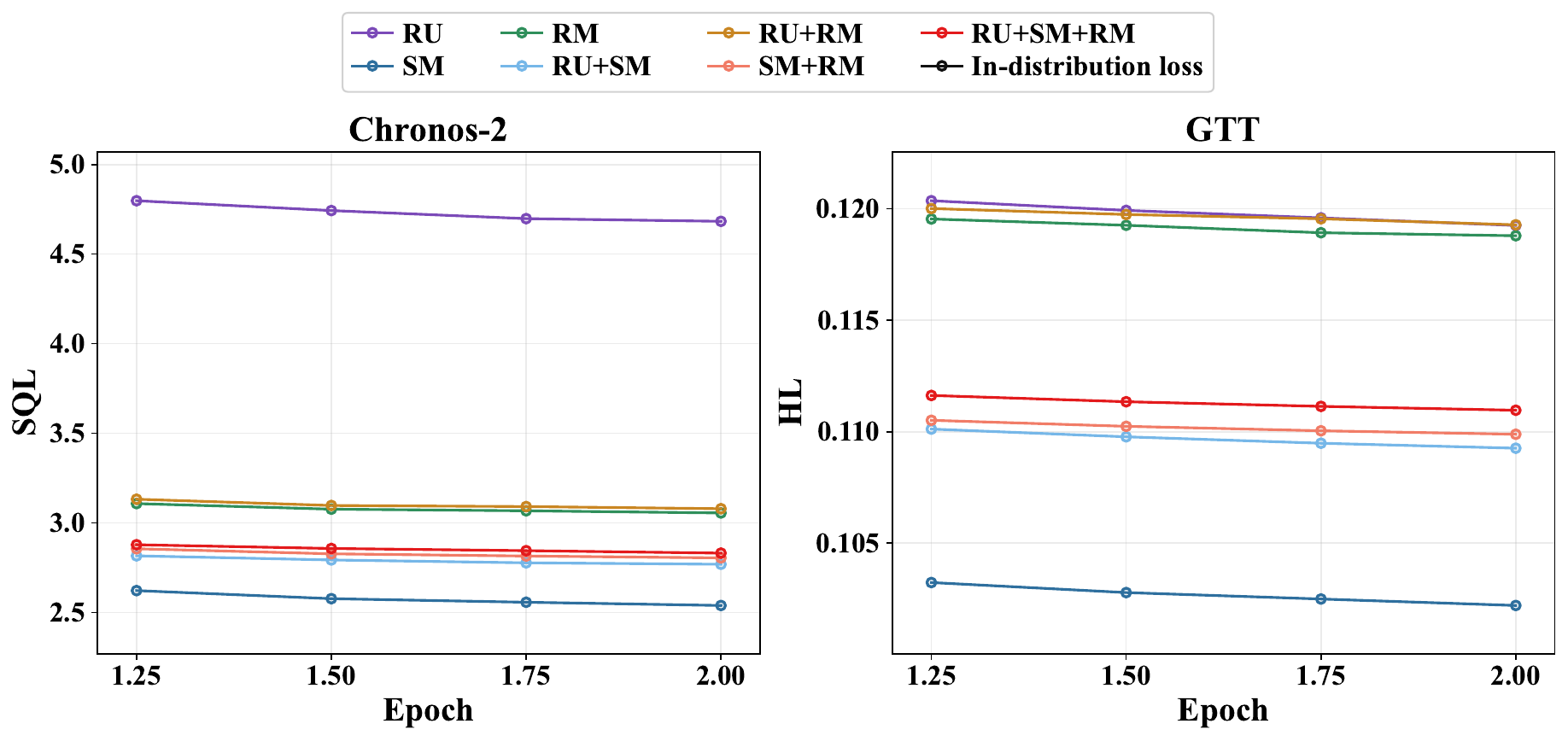}
    \caption{ID loss curves of Chronos-2 and GTT on different training corpora of the second epoch.}
    \label{fig:second_epoch_val}
\end{figure}

Figures~\ref{fig:first_epoch_train_val} and~\ref{fig:second_epoch_val} present the training and ID loss curves of four investigated TSFMs during pretraining. Both TimesFM-2.5 and Moirai-2.0 converge within $1$ epoch, as shown in Figure~\ref{fig:first_epoch_train_val}. In contrast, we observe that the training loss curves of both Chronos-2 and GTT indicate a downward trend within the first epoch. However, the ID loss curves become relatively stable by the end of the second epoch, as shown in Figure~\ref{fig:second_epoch_val}. Specifically, the ID loss curves of Chronos-2 decrease by less than 0.03 over the final quarter of the second epoch, while that of GTT decreases by less than 0.01. Moreover, we find that the second epoch does not consistently lead to better OOD benchmark performance than the first epoch, and even results in severe performance degradation in some cases, suggesting that additional training does not necessarily provide substantial OOD benefits, as detailed in Appendix~\ref{subapp:benchmark_mase}. Thus, we conclude that Chronos-2 and GTT also converge after 2 epochs.

Summing up the training dynamics of both univariate and multivariate models, we observe that the RM corpus does not achieve the lowest ID loss among the single-source corpora in most cases, which suggests that real-world multivariate data do not necessarily make pretraining easier in terms of convergence or ID loss. This may be associated with more complex real-world patterns and richer cross-variable dependencies in the RM corpus, making it more difficult to fit during pretraining.

\subsection{Out-of-Distribution Forecasting}\label{subsec:results}
\begin{figure}[t]
    \centering
    \includegraphics[width=0.95\linewidth]{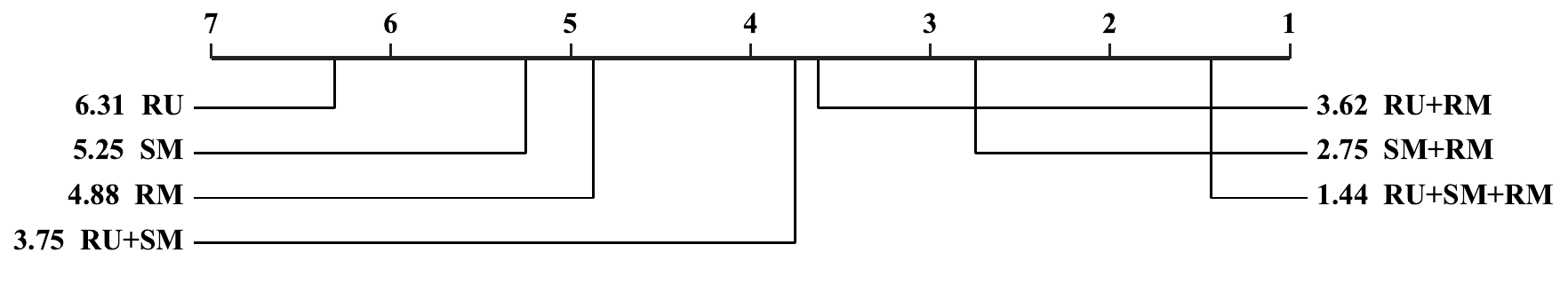}
    \caption{The average ranking related to MASE of seven corpora across all TSFMs and benchmarks.}
    \label{fig:ranking}
\end{figure}

\begin{figure}[h!]
    \centering

    \begin{subfigure}{0.495\linewidth}
        \centering
        \caption{Chronos-2}
        \label{subfig:chronos_results}
        \includegraphics[width=1\linewidth]{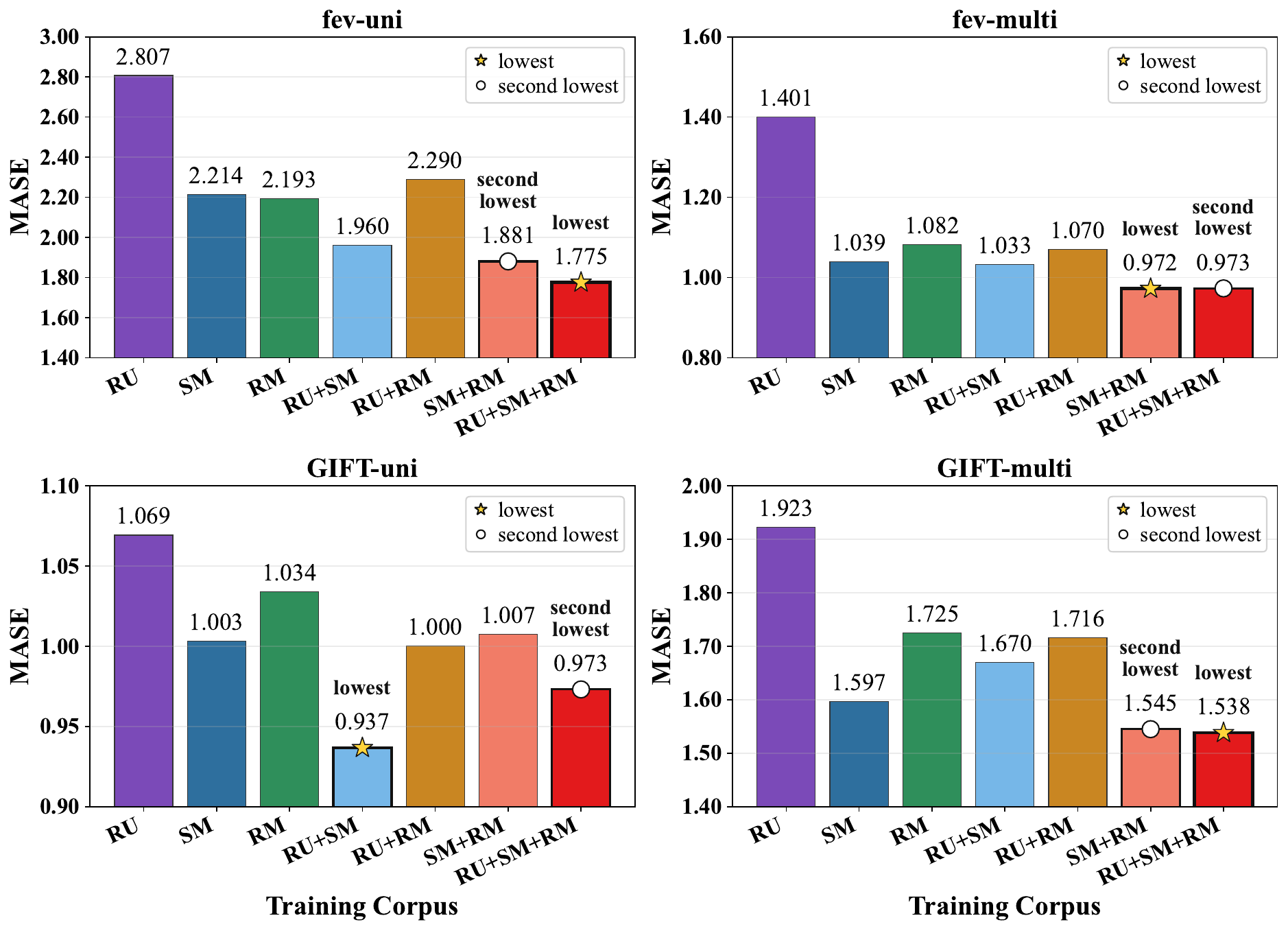}
    \end{subfigure}
    \hfill
    \begin{subfigure}{0.495\linewidth}
        \centering
        \caption{GTT}
        \label{subfig:gtt_results}
        \includegraphics[width=1\linewidth]{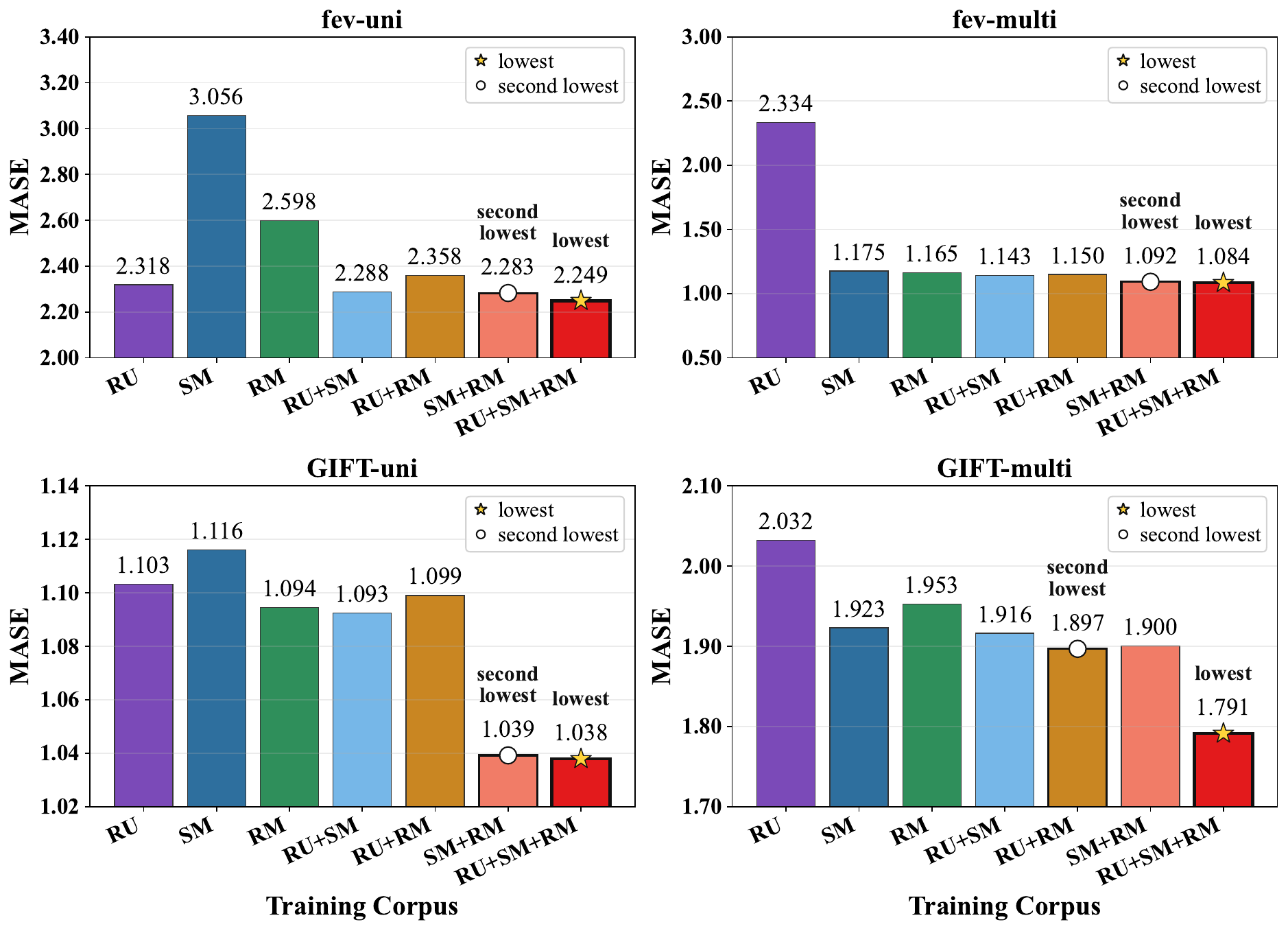}
    \end{subfigure}

    \vspace{0.3em}

    \begin{subfigure}{0.495\linewidth}
        \centering
         \caption{Moirai-2.0}
        \label{subfig:moirai_results}
        \includegraphics[width=1\linewidth]{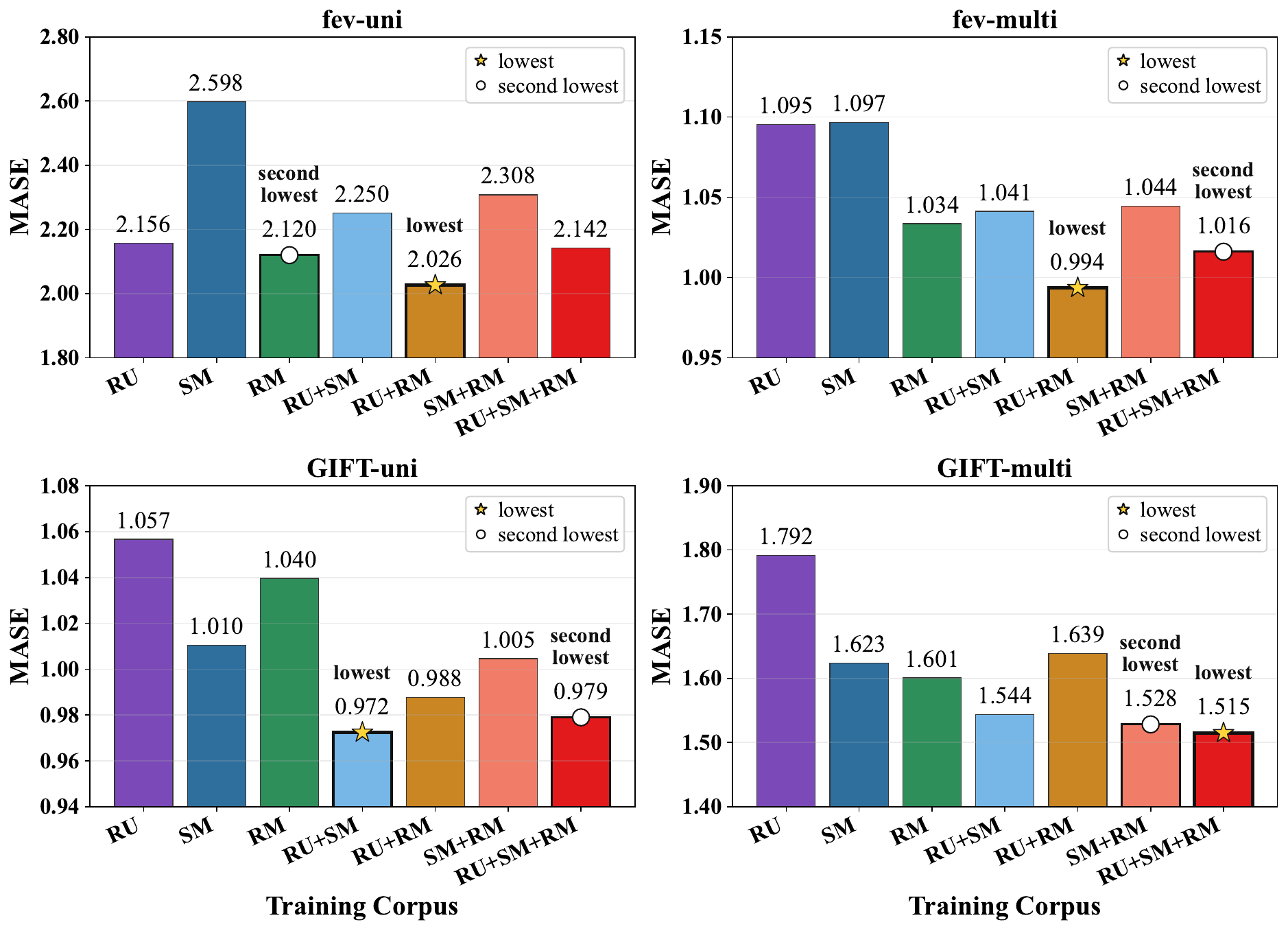}
    \end{subfigure}
    \hfill
    \begin{subfigure}{0.495\linewidth}
        \centering
        \caption{TimesFM-2.5}
        \label{subfig:timesfm_results}
        \includegraphics[width=1\linewidth]{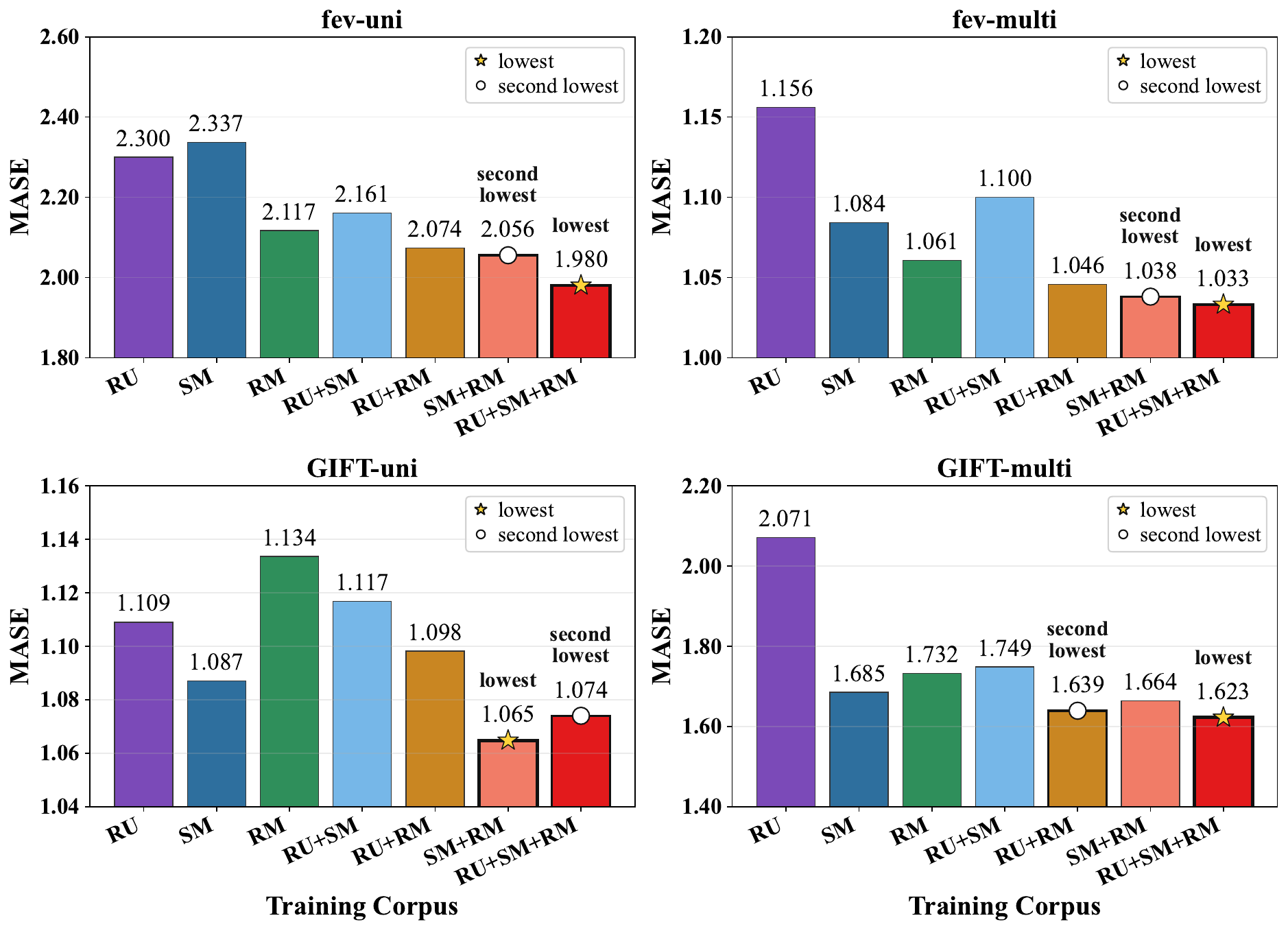}
    \end{subfigure}

    \caption{MASE results of different training corpora on OOD benchmarks for (a) Chronos-2, (b) GTT, (c) Moirai-2.0, and (d) TimesFM-2.5.}
    \label{fig:benchmark_results}
\end{figure}

This subsection provides an overall comparison of the OOD performance of TSFMs pretrained with different training corpora. To provide a unified OOD comparison across univariate and multivariate TSFMs, we evaluate all models on both the univariate and multivariate subsets of the benchmarks. For univariate TSFMs, evaluation on the multivariate subsets follows the same decomposition strategy as in Subsection~\ref{subsec:configurations} and processes each variable independently, without using covariates or cross-variable dependencies. We count the average ranks across seven training corpora according to the OOD MASE scores of four TSFMs. Figure~\ref{fig:ranking} reports the average ranks of seven corpora, where a lower rank indicates better forecasting performance that corresponds to the investigated corpus. It is observed that the RU corpus ranks last among all training corpora, suggesting the univariate dataset is less effective for pretraining TSFMs than the multivariate dataset. Furthermore, the RM corpus consistently outranks the SM corpus under both single-source-corpus and two-source-corpus pretraining. Specifically, the RM corpus achieves a better rank than the SM corpus, and the RU+RM corpus further outranks the RU+SM corpus. These results suggest that the real-world multivariate corpus is more advantageous for pretraining TSFMs than the synthetic multivariate corpus. Finally, the RU+SM+RM corpus achieves the best average rank, and the top three training corpora all contain the RM corpus. The comparisons indicate that incorporating a real-world multivariate corpus can further improve the performance of TSFMs.

We further examine the detailed results on each TSFM and OOD benchmark subset. Figure~\ref{fig:benchmark_results} displays the MASE results on all datasets and subset benchmarks for the TSFMs. We have the following observations and conclusions. (1) We observe that the RU corpus performs worse than the other training corpora in most cases, indicating that using the univariate dataset alone is insufficient to achieve strong OOD performance in our experiments. (2) The relative performance between the SM corpus and the RM corpus varies across models and benchmarks. For Chronos-2, the SM corpus outperforms the RM corpus, and the RU+SM corpus outperforms the RU+RM corpus. For GTT, the SM corpus outperforms the RM corpus in most cases, while the RU+SM corpus and the RU+RM corpus show comparable performance. For Moirai-2.0 and TimesFM-2.5, the RM corpus generally outperforms the SM corpus, and the RU+RM corpus outperforms the RU+SM corpus. These results indicate that neither the SM corpus nor the RM corpus is uniformly superior to the other, and both corpora have their respective strengths. (3) The RU+SM+RM corpus achieves the lowest MASE across most settings and reduces average MASE by 4.476\% compared with the currently widely used RU+SM corpus. Based on this, we recommend the RU+SM+RM corpus as the preferred pretraining corpus for building stronger multivariate TSFMs. Detailed MASE and WQL results for the OOD benchmarks can be accessed from Appendix~\ref{subapp:full_benchmark}.

\subsection{Case Studies}\label{subsec:case}
\begin{figure}[t]
    \centering

    \begin{subfigure}[t]{0.495\linewidth}
        \centering
        \caption{Electricity Price Forecasting Task}
        \label{subfig:case-price}
        \includegraphics[width=1\linewidth]{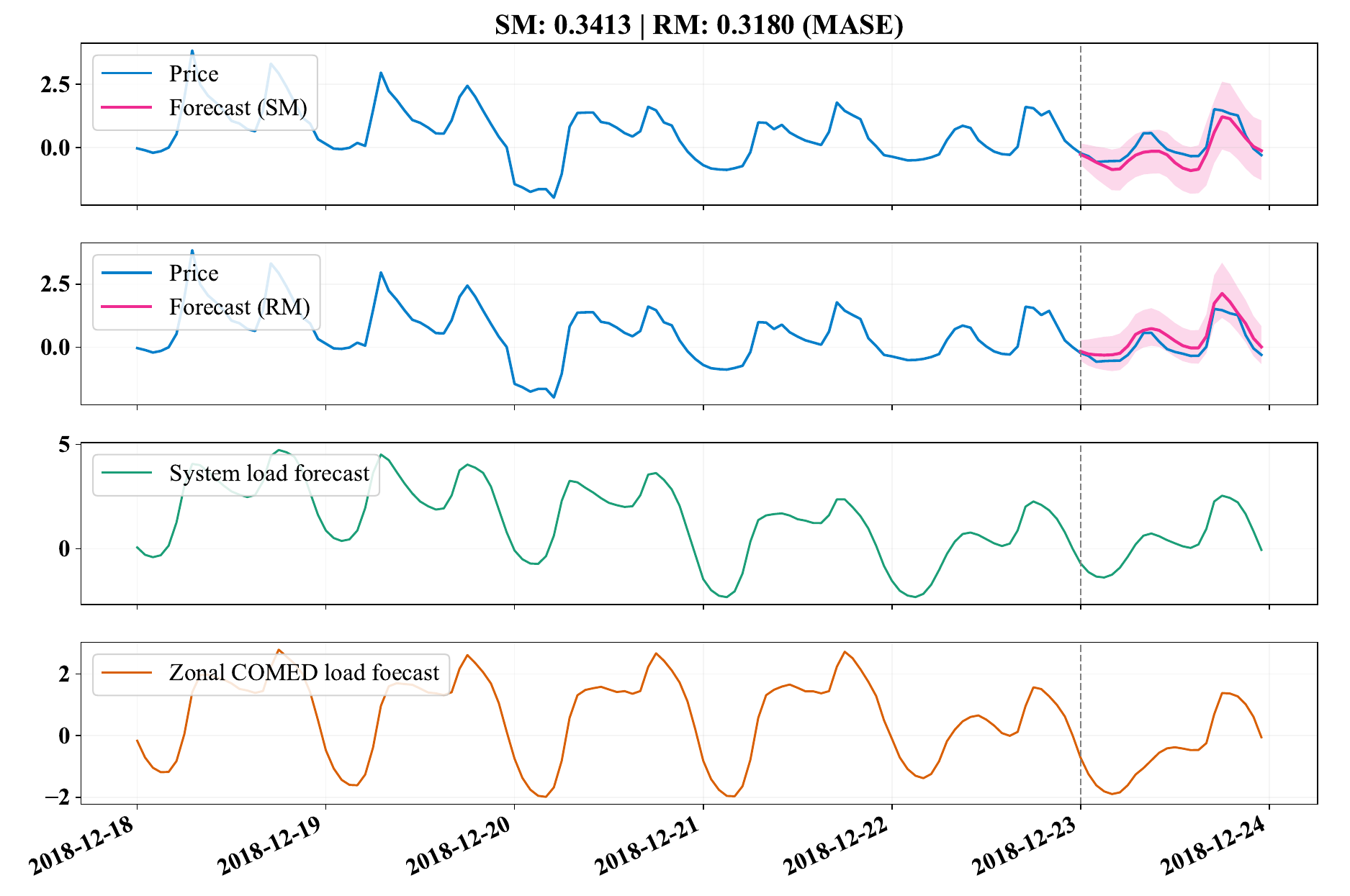}
    \end{subfigure}
    \hfill
    \begin{subfigure}[t]{0.495\linewidth}
        \centering
        \caption{Electricity Load Forecasting Task}
        \label{subfig:case-elec}
        \includegraphics[width=1\linewidth]{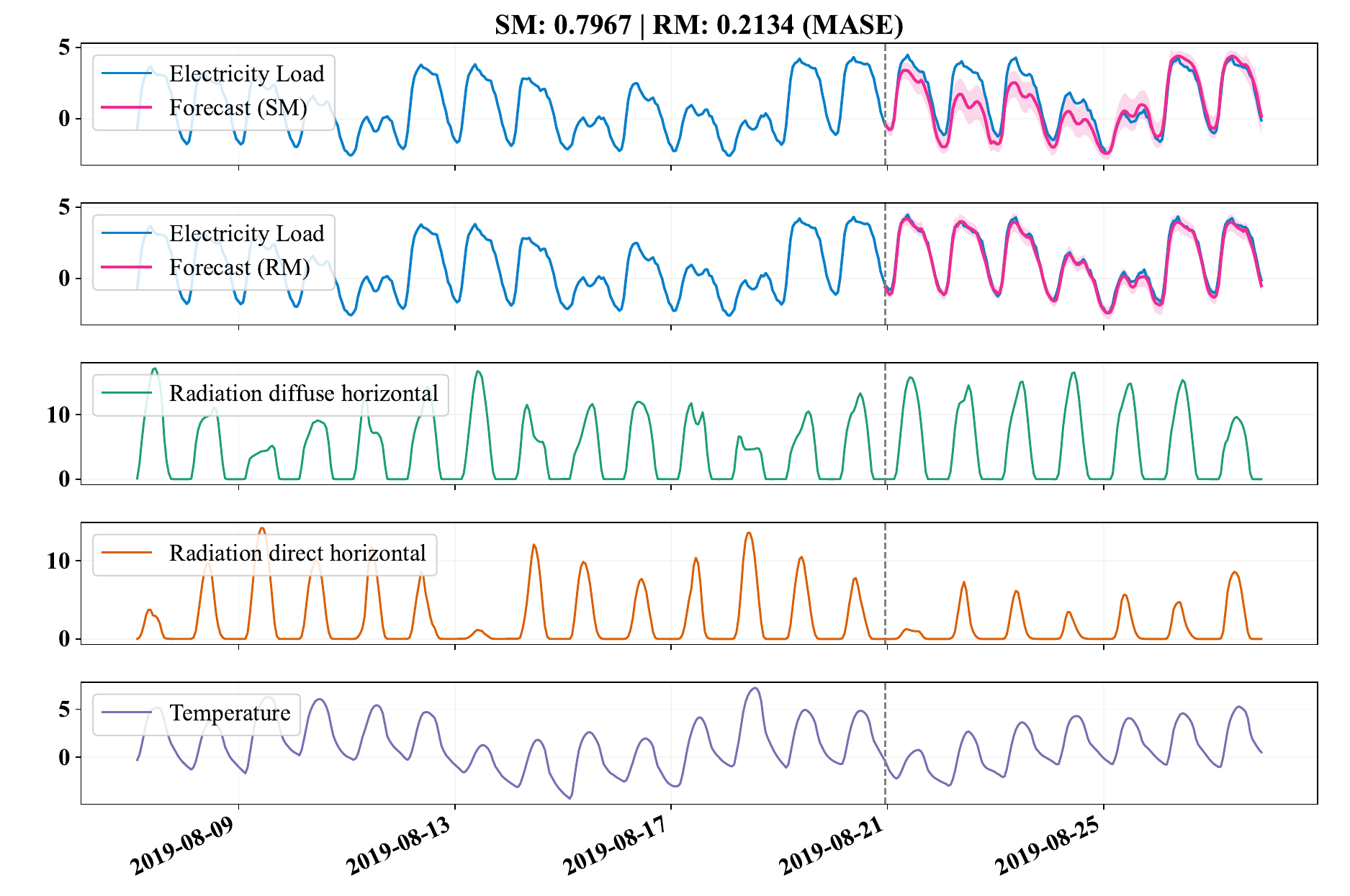}
    \end{subfigure}
    
    \caption{Forecasts generated by Chronos-2 models which are pretrained on the SM corpus and the RM corpus for two fev-bench tasks: (a) the electricity price forecasting task across the Pennsylvania, New Jersey, and Maryland zones and (b) the electricity load forecasting task from the ENTSO-E Transparency Platform. The forecasting horizon starts at the gray dashed vertical line, while the shaded area denotes the central 80\% prediction interval around the median forecast. For visualization, each target and covariate series is normalized, and the early part of the context window is truncated to improve visibility.}
    \label{fig:case}
\end{figure}

To visualize the advantage of real-world multivariate data in learning complex cross-variable dependencies, we compare the forecasts produced by the Chronos-2 model pretrained on the SM corpus and the RM corpus with respect to two representative samples, which cover simple and complex cross-variable dependencies, respectively. Figure~\ref{fig:case}(\subref{subfig:case-price}) shows a price forecasting task across the Pennsylvania, New Jersey, and Maryland zones. In this task, the next-day electricity price is forecasted using covariates of system load forecasts and zonal COMED load forecasts. The target and the two covariates exhibit highly similar trends, indicating that the cross-variable dependencies are relatively clear and easy to capture. For this case, the models pretrained on the SM corpus and on the RM corpus achieve comparable MASE, suggesting that both real-world multivariate data and synthetic multivariate data are sufficient for learning relatively simple cross-variable dependencies.

Figure~\ref{fig:case}(\subref{subfig:case-elec}) shows an hourly electricity load forecasting task from the ENTSO-E Transparency Platform. In this task, electricity load is forecasted using covariates of diffuse horizontal radiation, direct horizontal radiation, and temperature. Unlike the previous case, the target and covariates do not follow highly similar trends and show more complex cross-variable dependencies. In this setting, the model pretrained on the RM corpus obtains a much lower MASE than that pretrained on the SM corpus, with 0.2134 for RM and 0.7697 for SM. This suggests that real-world multivariate data provides stronger support for learning complex cross-variable dependencies and improves forecasting accuracy in such challenging settings.

\section{Conclusions}\label{sec:Discussions}
In this paper, we proposed the RMISC corpus for supporting pretraining and benchmarking TSFMs with large-scale and real-world multivariate time series. In systematic comparisons with real-world univariate and synthetic multivariate corpora, we confirmed that the RMISC corpus provides valuable multivariate information from realistic contexts and can effectively complement existing pretraining data. In particular, the combination of real-world univariate data, synthetic multivariate data, and RMISC leads to more robust zero-shot generalization than the currently widely used pretraining corpus. These results suggest that our proposed RMISC corpus provides an effective data foundation for building multivariate TSFMs.

\section*{Acknowledgements}
This research was supported by the Nanjing University-Siemens Joint Research Center for Industrial AI, Jiangsu Science and Technology Project (BG2024031).

\bibliography{JMref}
\bibliographystyle{unsrtnat}

\clearpage

\appendix

{\Large\bfseries Appendix}

This appendix provides the supplementary materials for our work 
``RMISC: A Large-scale Real-world Multivariate Corpus for Time Series Foundation Models''.

\section{Characteristics of the RMISC Corpus}\label{app:dataset}

\subsection{Full properties of the RMISC Corpus}\label{app:full_table}

Table~\ref{tab:rmisc_datasets} summarizes the datasets and key properties of RMISC, including domain, frequency, dimensionality, scale, and original source.

	{\scriptsize
	\begin{longtable}{|c|c|c|c|r|r|c|}
		\caption{Datasets and key properties of RMISC. ``Freq." denotes the sampling frequency  (ms = millisecond, s = second, min = minute, h = hour, d = day, m = month, y = year, ``-" signifies multiple values or unknown frequency); ``Dim." represents the average dimension of the dataset; ``Time Steps" represents the number of time steps within the dataset; ``Obs." refers to the total count of time points; and ``Source" denotes the original paper or resource of the dataset.} \label{tab:rmisc_datasets} \\
		\hline
		\textbf{Dataset} & \textbf{Domain} & \textbf{Freq.} & \textbf{Dim.} & \textbf{Time Steps} & \textbf{Obs.} & \textbf{Source} \\
		\hline\hline
		\endfirsthead
		
		\multicolumn{7}{c}{{\bfseries \tablename\ \thetable{} -- continued from previous page}} \\
		\hline
		\textbf{Dataset} & \textbf{Domain} & \textbf{Freq.} & \textbf{Dim.} & \textbf{Time Steps} & \textbf{Obs.} & \textbf{Source} \\
		\hline
		\endhead
		
		\hline
		\multicolumn{7}{|r|}{{Continued on next page}} \\
		\hline
		\endfoot
		
		\hline
		\endlastfoot
		
		ACSF1 & Energy & - & 1 & 0.29 M & 0.29 M & ~\cite{gisler2013acsd,schafer2017weasel} \\
		ApplianceEnergy & Energy & 10min & 26 & 0.02 M & 0.51 M & ~\cite{candanedo2017appliances} \\
		AustralianElectricityDemand & Energy & 30min & 5 & 0.23 M & 1.15 M & ~\cite{godahewa2021australian} \\
		AzurePublicDatasetV1 & Energy & 5min & 3 & 1020.38 M & 3060.08 M & ~\cite{cortez2017resource} \\
		AzurePublicDatasetV2 & Energy & 5min & 3 & 1656.28 M & 4968.71 M & ~\cite{cortez2017resource} \\
		BDG2-Bear & Energy & h & 1 & 1.42 M & 1.42 M & ~\cite{miller2020building,xiaoming2025time} \\
		BDG2-Fox & Energy & h & 1 & 2.29 M & 2.29 M & ~\cite{miller2020building,xiaoming2025time} \\
		BDG2-Panther & Energy & h & 1 & 0.89 M & 0.89 M & ~\cite{miller2020building,xiaoming2025time} \\
		BDG2-Rat & Energy & h & 1 & 4.60 M & 4.60 M & ~\cite{miller2020building,xiaoming2025time} \\
		BatteryRUL & Energy & - & 9 & 0.02 M & 0.14 M & Kaggle\footnote{https://www.kaggle.com/} \\
		BritainCoal & Energy & - & 10 & 0.80 M & 7.96 M & Data.World\footnote{https://data.world/} \\
		BuildingsBenchComAmy & Energy & h & 148 & 20.59 M & 3040.60 M & ~\cite{Emami2023BuildingsBench} \\
		BuildingsBenchComTmy & Energy & h & 147 & 20.59 M & 3026.98 M & ~\cite{Emami2023BuildingsBench} \\
		BuildingsBenchRealCSV & Energy & h & 2 & 22.25 M & 39.64 M & ~\cite{Emami2023BuildingsBench} \\
		BuildingsBenchResAmy & Energy & h & 235 & 20.46 M & 4815.70 M & ~\cite{Emami2023BuildingsBench} \\
		BuildingsBenchResTmy & Energy & h & 235 & 20.46 M & 4815.72 M & ~\cite{Emami2023BuildingsBench} \\
		Bull & Energy & - & 1 & 0.50 M & 0.50 M & ~\cite{xiaoming2025time} \\
		Computers & Energy & 2min & 1 & 0.36 M & 0.36 M & TSC.\footnote{https://www.timeseriesclassification.com/} \\
		ERCOT & Energy & h & 8 & 0.17 M & 1.39 M & ERCOT\footnote{https://www.ercot.com/} \\
		ETT & Energy & 15min & 7 & 0.17 M & 1.22 M & ~\cite{haoyietal-informer-2021} \\
		ETTMulti & Energy & 15min & 7 & 0.17 M & 1.22 M & ~\cite{haoyietal-informer-2021} \\
		Electricity & Energy & - & 321 & 0.03 M & 8.44 M & ~\cite{trindade_2015_electricity} \\
		ElectricityHourly & Energy & h & 321 & 0.03 M & 8.44 M & ~\cite{godahewa_2020_electricity} \\
		GFC2012 & Energy & h & 1 & 0.50 M & 0.50 M & ~\cite{xiaoming2025time, wang2023benchmarks} \\
		Hog & Energy & h & 1 & 0.37 M & 0.37 M & ~\cite{woo2024unified,xiaoming2025time} \\
		HouseholdPower & Energy & h & 7 & 2.08 M & 14.53 M & ~\cite{individual_household_electric_power_consumption_235} \\
		Ideal & Energy & h & 1 & 1.25 M & 1.25 M & ~\cite{woo2024unified,xiaoming2025time} \\
		LondonSmartMeters & Energy & 30min & 1 & 71.93 M & 71.93 M & ~\cite{godahewa_2020_4656072} \\
		OPSD & Energy & h & 8 & 2.86 M & 22.90 M & OPSD\footnote{https://data.open-power-system-data.org/} \\
		OPSD-Household & Energy & 15min & 7 & 7.12 M & 47.88 M & OPSD \\
		OPSD-PV-Wind & Energy & h & 4 & 12.62 M & 48.74 M & ~\cite{Pfenninger2016PV,Staffell2016Wind} \\
		OPSD-When2Heat & Energy & h & 22 & 2.09 M & 45.61 M & ~\cite{ruhnau2019heatdemand} \\
		OilWell & Energy & - & 5 & 50.91 M & 244.53 M & ~\cite{vargas2019realistic} \\
		Pvdaq & Energy & 15min & 2 & 3.97 M & 8.21 M & OEDI\footnote{https://data.openei.org/} \\
		ResidentialPower & Energy & min & 2 & 262.85 M & 525.09 M & ~\cite{bergmeir_2023_8219786} \\
		ShellHackathon & Energy & - & 15 & 0.53 M & 7.91 M & Kaggle \\
		Solar10Minutes & Energy & 10min & 137 & 0.05 M & 7.20 M & ~\cite{godahewa_2020_solar10min} \\
		Solar4Seconds & Energy & 4s & 1 & 7.40 M & 7.40 M & ~\cite{godahewa_2020_solar} \\
		SolarEnergy & Energy & 10min & 137 & 0.05 M & 7.20 M & ~\cite{lai2018modeling} \\
		TetuanPowerConsumption & Energy & 10min & 8 & 0.05 M & 0.42 M & ~\cite{salam_2018_tetouan_power} \\
		UK-DALE & Energy & - & 2 & 30.20 M & 65.60 M & ~\cite{kelly2015ukdale} \\
		WindElec & Energy & 15min & 13 & 0.23 M & 3.01 M & DCIC\footnote{https://www.dcic-china.com/competitions/10098/datasets} \\
		WindFarms & Energy & min & 295 & 0.07 M & 19.26 M & ~\cite{godahewa2020windfarms} \\
		WindPower4secs & Energy & 4s & 1 & 7.40 M & 7.40 M & ~\cite{godahewa2020windpower} \\
		BeijingAirQuality & Environment & h & 8 & 0.42 M & 3.16 M & ~\cite{chen2017beijing} \\
		BeutenbergWeather & Environment & - & 20 & 0.89 M & 17.88 M & Kaggle \\
		CMIP6-2000-PartI & Environment & 6h & 1 & 1056.50 M & 1056.50 M & ~\cite{xiaoming2025time,Eyring2016CMIP6} \\
		CMIP6-2000-PartII & Environment & 6h & 1 & 1056.50 M & 1056.50 M & ~\cite{xiaoming2025time,Eyring2016CMIP6} \\
		CMIP6-2000-PartIII & Environment & 6h & 1 & 1056.49 M & 1056.49 M & ~\cite{xiaoming2025time,Eyring2016CMIP6} \\
		CMIP6-2005-PartI & Environment & 6h & 1 & 1056.50 M & 1056.50 M & ~\cite{xiaoming2025time,Eyring2016CMIP6} \\
		CMIP6-2005-PartII & Environment & 6h & 1 & 1056.50 M & 1056.50 M & ~\cite{xiaoming2025time,Eyring2016CMIP6} \\
		CMIP6-2005-PartIII & Environment & 6h & 1 & 1056.49 M & 1056.49 M & ~\cite{xiaoming2025time,Eyring2016CMIP6} \\
		CMIP6-2010-PartI & Environment & 6h & 1 & 1056.50 M & 1056.50 M & ~\cite{xiaoming2025time,Eyring2016CMIP6} \\
		CMIP6-2010-PartII & Environment & 6h & 1 & 1056.50 M & 1056.50 M & ~\cite{xiaoming2025time,Eyring2016CMIP6} \\
		CMIP6-2010-PartIII & Environment & 6h & 1 & 1056.49 M & 1056.49 M & ~\cite{xiaoming2025time,Eyring2016CMIP6} \\
		ERA5HourlySingleLevels & Environment & h & 15 & 30.86 M & 462.92 M & ~\cite{Nguyen2023ClimateLearn} \\
		GasSensorTemperature & Environment & - & 20 & 3.84 M & 76.86 M & ~\cite{gas_sensor_array_temperature_modulation_487} \\
		GlobalClimateChange & Environment & m & 2 & 2.81 M & 5.63 M & Data.World \\
		KDDCup2018 & Environment & h & 50 & 0.01 M & 0.54 M & ~\cite{godahewa_2020_4656719} \\
		OikolabWeather & Environment & h & 8 & 0.10 M & 0.80 M & ~\cite{godahewa_2021_5184708} \\
		PM25FiveCities & Environment & h & 10 & 0.11 M & 1.15 M & ~\cite{chen2016pm25} \\
		Subseasonal & Environment & d & 60 & 93.79 M & 5668.67 M & ~\cite{mouatadid2023subseasonal} \\
		TemperatureRain & Environment & d & 1614 & 0.0007 M & 1.17 M & ~\cite{godahewa_2021_temperature_rain} \\
		Tigge & Environment & 6h & 194 & 0.11 M & 21.01 M & ~\cite{Rasp_2020} \\
		USAirPollution & Environment & - & 14 & 1.75 M & 24.45 M & Data.World \\
		Weather & Environment & d & 1 & 14.72 M & 14.72 M & ~\cite{godahewa2020weather} \\
		WeatherBench5-625deg & Environment & h & 61 & 684.84375 M & 43783.91 M & ~\cite{Rasp_2020} \\
		WeatherTest & Environment & - & 21 & 0.05 M & 1.11 M & MPIB\footnote{https://www.bgc-jena.mpg.de/wetter/} \\
		XiamenAirQuality & Environment & h & 6 & 1.54 M & 9.10 M & DataCastle\footnote{https://challenge.datacastle.cn/v3/cmptDetail.html?id=950} \\
		AMarketChina & Finance & - & 6 & 0.62 M & 3.71 M & ~\cite{wu2022price} \\
		AMarketChinaKnownOpen & Finance & - & 6 & 0.62 M & 3.71 M & ~\cite{wu2022price} \\
		AliCar & Finance & - & 2 & 0.005 M & 0.01 M & Aliyun\footnote{https://tianchi.aliyun.com/competition/entrance/231641/information} \\
		Bitcoin & Finance & d & 645 & 0.004 M & 2.83 M & Kaggle \\
		Bizitobs\_application & Finance & 10s & 1 & 0.01 M & 0.02 M & ~\cite{aksu2024gift,palaskar2024automixer} \\
		Bizitobs\_l2c\_H & Finance & h & 1 & 0.00 M & 0.02 M & ~\cite{aksu2024gift,palaskar2024automixer} \\
		CSI500 & Finance & min & 7 & 91.96 M & 643.70 M & CSI\footnote{https://www.csindex.com.cn} \\
		CausalEffects & Finance & - & 100 & 0.001 M & 0.11 M & Data.World \\
		ChinaMinuteStock & Finance & min & 13 & 498.52 M & 6480.79 M & Hugging Face\footnote{https://huggingface.co/} \\
		Cif2016-12 & Finance & m & 1 & 0.006 M & 0.006 M & ~\cite{woo2024unified,xiaoming2025time} \\
		Cif2016-6 & Finance & m & 1 & 0.0006 M & 0.0006 M & ~\cite{woo2024unified,xiaoming2025time} \\
		Cryptocurrency & Finance & - & 5 & 1.97 M & 9.87 M & Kaggle \\
		CryptocurrencyKnownOpen & Finance & - & 5 & 1.97 M & 9.87 M & Kaggle \\
		Dominick & Finance & - & 1298 & 0.0004 M & 0.51 M & ~\cite{godahewa_2020_4654802} \\
		ExchangeRate & Finance & - & 8 & 0.01 M & 0.06 M & ~\cite{lai2018modeling} \\
		FavoritaSales & Finance & d & 28 & 15.85 M & 448.49 M & ~\cite{favorita-grocery-sales-forecasting} \\
		FavoritaTransactions & Finance & d & 3 & 0.08 M & 0.25 M & ~\cite{favorita-grocery-sales-forecasting} \\
		FavoritaTransactionsKnownOil & Finance & d & 3 & 0.08 M & 0.25 M & ~\cite{favorita-grocery-sales-forecasting} \\
		FredMD & Finance & m & 110 & 0.0007 M & 0.08 M & ~\cite{godahewa_2020_4654833} \\
		HierachicalSales & Finance & d & 234 & 0.002 M & 0.42 M & ~\cite{mancuso2021hierarchical} \\
		KaggleTS & Finance & - & 6 & 0.01 M & 0.05 M & Kaggle \\
		M5 & Finance & - & 318 & 0.37 M & 116.21 M & ~\cite{m5-forecasting-accuracy} \\
		NIFTYStock & Finance & - & 9 & 0.47 M & 4.24 M & Kaggle \\
		NIFTYStockKnownOpen & Finance & - & 9 & 0.47 M & 4.24 M & Kaggle \\
		NN5Daily & Finance & d & 114 & 0.0008 M & 0.09 M & ~\cite{godahewa_2020_4656110} \\
		Restaurant & Finance & - & 1 & 0.03 M & 0.03 M & ~\cite{xiaoming2025time} \\
		Rohlik\_orders\_1D & Finance & d & 7 & 0.01 M & 0.01 M & ~\cite{shchur2025fev} \\
		Rohlik\_orders\_1W & Finance & w & 7 & 0.00 M & 0.00 M & ~\cite{shchur2025fev} \\
		Rossmann\_1D & Finance & d & 1115 & 1.05 M & 1.05 M & ~\cite{shchur2025fev} \\
		Rossmann\_1W & Finance & w & 1115 & 0.15 M & 0.15 M & ~\cite{shchur2025fev} \\
		SP500 & Finance & - & 5 & 0.60 M & 3.01 M & ~\cite{sidi2020improving} \\
		SP500KnownOpen & Finance & - & 5 & 0.60 M & 3.01 M & ~\cite{sidi2020improving} \\
		StockFactorsCleaned & Finance & m & 70 & 16.20 M & 1133.71 M & Hugging Face \\
		StockMarketData & Finance & - & 70 & 0.01 M & 0.69 M & Kaggle \\
		TourismMonthly & Finance & m & 1 & 0.10 M & 0.10 M & ~\cite{xiaoming2025time} \\
		TushareETFDaily & Finance & d & 10 & 2.44 M & 24.36 M & Tushare\footnote{https://tushare.pro/} \\
		TushareIndexDaily & Finance & d & 11 & 2.64 M & 26.40 M & Tushare \\
		TushareStockDaily & Finance & d & 11 & 14.16 M & 155.79 M & Tushare \\
		TushareStockDailyMetrics & Finance & d & 14 & 14.03 M & 196.43 M & Tushare \\
		TushareStockWeekly & Finance & w & 11 & 2.97 M & 32.64 M & Tushare \\
		UKEconomy & Finance & - & 1 & 0.39 M & 0.40 M & Data.World \\
		WeeklyFuelPricesItaly & Finance & w & 4 & 0.005 M & 0.02 M & Data.World \\
		WeeklyRoadFuelPrices & Finance & w & 2 & 0.0009 M & 0.002 M & Data.World \\
		BTS & Industry & - & 1495 & 0.06 M & 95.87 M & ~\cite{prabowo2024bts} \\
		Behavior-1k & Industry & - & 446 & 84.40 M & 37682.52 M & ~\cite{li2024behavior} \\
		FrothFlotation & Industry & - & 12 & 0.003 M & 0.04 M & Kaggle \\
		GasPipeline & Industry & - & 10 & 0.14 M & 1.38 M & ~\cite{beaver2013scada} \\
		GasSensorDynamic & Industry & - & 18 & 2.10 M & 37.75 M & ~\cite{gas_sensor_array_under_dynamic_gas_mixtures_322} \\
		LBNL & Industry & min & 61 & 1.99 M & 122.27 M & ~\cite{hong2022three_year_building} \\
		OccupancyDetection & Industry & - & 6 & 0.02 M & 0.12 M & ~\cite{occupancy_detection__357} \\
		PUMP & Industry & - & 44 & 0.22 M & 9.69 M & Kaggle \\
		ProEnFo & Industry & h & 23 & 0.23 M & 5.31 M & ~\cite{wang2023benchmarks} \\
		RoomOccupancy & Industry & 30s & 17 & 0.01 M & 0.17 M & ~\cite{room_occupancy_estimation_864} \\
		SWAT & Industry & 5s & 42 & 0.19 M & 7.93 M & ~\cite{goh_2017_securewater} \\
		ServerMachineDataset & Industry & - & 31 & 0.71 M & 21.99 M & ~\cite{su2019robust} \\
		SmellSensor & Industry & m & 19 & 21.19 M & 402.56 M & Hugging Face \\
		WADI & Industry & 5s & 93 & 0.26 M & 23.96 M & Kaggle \\
		BeijingSubway & Traffic & 30min & 276 & 0.01 M & 2.98 M & ~\cite{zhang2020deep} \\
		ChengduTaxi & Traffic & - & 4 & 0.71 M & 2.85 M & ~\cite{wang2018will} \\
		LoopSeattleLA & Traffic & 5min & 258 & 0.06 M & 15.89 M & ~\cite{10.1145/3474717.3483923} \\
		Mdense & Traffic & - & 1 & 0.02 M & 0.02 M & ~\cite{de2020spatio} \\
		Metropt3 & Traffic & - & 15 & 1.05 M & 15.73 M & ~\cite{metropt-3_dataset_791} \\
		MetroTraffic & Traffic & - & 5 & 0.05 M & 0.24 M & ~\cite{metro_interstate_traffic_volume_492} \\
		PEMS-Bay-METRO-LA & Traffic & 5min & 278 & 0.09 M & 24.03 M & ~\cite{li2018dcrnn} \\
		PEMSCalifornia & Traffic & - & 361 & 0.11 M & 38.22 M & ~\cite{10.1145/3474717.3483923} \\
		QtrafficSpeed & Traffic & - & 2 & 264.39 M & 528.77 M & ~\cite{bbliaojqZhangKDD18deep} \\
		Rideshare & Traffic & h & 1969 & 0.0002 M & 0.38 M & ~\cite{godahewa_2021_5122232} \\
		SHandHZMetro & Traffic & 15min & 241 & 0.08 M & 20.38 M & ~\cite{10.1145/3474717.3483923} \\
		T-Drive & Traffic & 10min & 3 & 17.66 M & 52.99 M & ~\cite{Yuan2011Driving,Yuan2010Tdrive} \\
		Traffic & Traffic & h & 862 & 0.02 M & 15.12 M & ~\cite{lai2018modeling} \\
		TrafficHourly & Traffic & h & 862 & 0.02 M & 15.12 M & ~\cite{godahewa2020_traffic_hourly} \\
		WikiTrafficDaily & Traffic & d & 1 & 304.48 M & 304.48 M & ~\cite{godahewa2020kagglewiki} \\
		WikiTrafficWeekly & Traffic & w & 1 & 16.39 M & 16.39 M & ~\cite{godahewa2020kagglewikiweekly} \\
		BCI\_Competetion\_IV\_1 & Others & 10ms & 59 & 3.01 M & 177.37 M & ~\cite{blankertz2007noninvasive} \\
		BCI\_Competetion\_IV\_2a & Others & 4ms & 19 & 7.39 M & 143.09 M & ~\cite{brunner2008bci} \\
		BCI\_Competetion\_IV\_2b & Others & 4ms & 3 & 8.46 M & 25.37 M & ~\cite{leeb2008bci} \\
		BooksPerPerson & Others & - & 1 & 0.01 M & 0.01 M & Data.World \\
		BoschCNC & Others & 500us & 3 & 34.07 M & 102.20 M & ~\cite{tnani2022smart} \\
		BrainInvadersBi2014b & Others & 2ms & 33 & 17.39 M & 573.90 M & ~\cite{korczowski_2019_3267302} \\
		CSE-CIC-IDS2018 & Others & - & 78 & 16.23 M & 1266.17 M & ~\cite{Sharafaldin2018ICISSP} \\
		CSTSNonnormalTest & Others & s & 4 & 37.96 M & 151.83 M & ~\cite{degen2025csts} \\
		CSTSNonnormalTrain & Others & s & 4 & 37.92 M & 151.68 M & ~\cite{degen2025csts} \\
		CSTSNormalTest & Others & s & 4 & 37.96 M & 151.83 M & ~\cite{degen2025csts} \\
		CSTSNormalTrain & Others & s & 4 & 37.92 M & 151.68 M & ~\cite{degen2025csts} \\
		Car & Others & - & 1 & 0.07 M & 0.07 M & ~\cite{thakoor2005shape} \\
		CinCECGTorso & Others & - & 1 & 2.33 M & 2.33 M & TSC. \\
		Covid & Others & - & 7 & 0.001 M & 0.01 M & ~\cite{hasell2020covid_testing,mathieu2021covid_vaccinations} \\
		CovidDeaths & Others & - & 236 & 0.0002 M & 0.05 M & ~\cite{godahewa2020covid_deaths} \\
		CovidMobility & Others & d & 218 & 0.0004 M & 0.09 M & ~\cite{godahewa2021covid_mobility} \\
		Darts & Others & - & 15 & 0.05 M & 0.71 M & Darts\footnote{https://unit8co.github.io/darts/} \\
		EMG4Gestures & Others & - & 9 & 4.24 M & 38.14 M & ~\cite{krilova_2018_emg} \\
		EbayServer & Others & - & 26 & 0.13 M & 3.44 M & ~\cite{abdulaal_2021_rt_sync,abdulaal_2021_async} \\
		EigenWorms & Others & - & 6 & 4.66 M & 27.95 M & ~\cite{yemini_2013_celegans} \\
		FordA & Others & - & 1 & 2.46 M & 2.46 M & TSC. \\
		Gait & Others & - & 7 & 0.18 M & 1.27 M & ~\cite{multivariate_gait_data_760} \\
		HAR70Plus & Others & - & 7 & 2.26 M & 15.82 M & ~\cite{har70+_780} \\
		HARTH & Others & - & 7 & 3.96 M & 27.75 M & ~\cite{harth_779} \\
		HetergeneousHAR & Others & 5ms & 7 & 14.13 M & 98.90 M & ~\cite{heterogeneity_activity_recognition_344} \\
		HungarianChickenpoxCases & Others & - & 19 & 0.0005 M & 0.01 M & ~\cite{hungarian_chickenpox_cases_580} \\
		Illness & Others & w & 10 & 0.001 M & 0.01 M & FluView\footnote{https://gis.cdc.gov/grasp/fluview/} \\
		IndoorLocalisation & Others & 100ms & 12 & 0.15 M & 1.88 M & ~\cite{barsocchi2016multisource} \\
		InlineSkate & Others & - & 1 & 1.22 M & 1.22 M & ~\cite{morchen2006time} \\
		KeplerLightCurves & Others & - & 1 & 5.89 M & 5.89 M & ~\cite{barbara2022classifying} \\
		LargeST & Others & 5min & 125 & 35.66 M & 4439.10 M & Kaggle \\
		M3 & Others & - & 1 & 0.23 M & 0.23 M & Monash TSF.\footnote{https://forecastingdata.org/} \\
		M4 & Others & - & 1 & 19.65 M & 19.65 M & Monash TSF. \\
		MelbournePedestrianCounts & Others & h & 1 & 3.13 M & 3.13 M & ~\cite{godahewa_2020_4656626} \\
		MiniApp & Others & - & 26 & 0.01 M & 0.34 M & ~\cite{li2024functional} \\
		MotionSense & Others & - & 3 & 2.47 M & 7.42 M & ~\cite{Malekzadeh:2019:MSD:3302505.3310068} \\
		MotorTemperature & Others & - & 12 & 1.33 M & 15.97 M & ~\cite{wilhelm_kirchgassner_2021} \\
		MZVAV & Others & - & 17 & 0.40 M & 6.83 M & ~\cite{Granderson2020} \\
		NAB & Others & - & 1 & 0.32 M & 0.32 M & ~\cite{ahmad2017unsupervised} \\
		PAMAP2 & Others & 10ms & 41 & 2.72 M & 111.72 M & ~\cite{reiss2012pamap2} \\
		Rebound & Others & - & 6001 & 0.02 M & 120.02 M & Hugging Face \\
		Satellite & Others & - & 15 & 0.19 M & 2.91 M & ~\cite{hundman_2018_spacecraft} \\
		SmartMeterAus30m & Others & 30min & 2 & 344.74 M & 1034.22 M & Hugging Face \\
		SmartMeterAus60m & Others & h & 2 & 172.97 M & 345.93 M & Hugging Face \\
		SmartMeterUK30m & Others & 30min & 2 & 166.88 M & 500.65 M & Hugging Face \\
		SmartMeterUK60m & Others & h & 2 & 83.81 M & 167.62 M & Hugging Face \\
		StarLightCurves & Others & - & 9.24 K & 1.02 K & 9.46 M & ~\cite{keogh2006lb_keogh} \\
		Sunspots & Others & - & 1 & 0.003 M & 0.003 M & Kaggle \\
		TimeMMD & Others & - & 3 & 0.02 M & 0.10 M & ~\cite{liu2024timemmd} \\
		USBirths & Others & - & 1 & 0.01 M & 0.01 M & ~\cite{godahewa2020usbirths} \\
		VehicleTrips & Others & d & 4 & 0.0002 M & 0.0008 M & ~\cite{godahewa2021vehicletrips} \\
		WISDM\_V1 & Others & 50ms & 4 & 0.99 M & 3.95 M & ~\cite{kwapisz2010activity,weiss2012personalization} \\
		WISDM\_V2 & Others & 50ms & 4 & 2.69 M & 10.25 M & ~\cite{kwapisz2010activity,weiss2012personalization} \\
		WISDM\_V3 & Others & 50ms & 13 & 2.99 M & 38.88 M & ~\cite{kwapisz2010activity,weiss2012personalization} \\
		Worms & Others & - & 1 & 0.23 M & 0.23 M & TSC. \\
		
	\end{longtable}

}

\subsection{Statistical Analyses}\label{app:statistics}
This subsection provides additional statistical analyses of the proposed RMISC corpus, further demonstrating its scale, diversity, and quality.

\begin{figure}[t]
    \centering
    \includegraphics[width=1\linewidth]{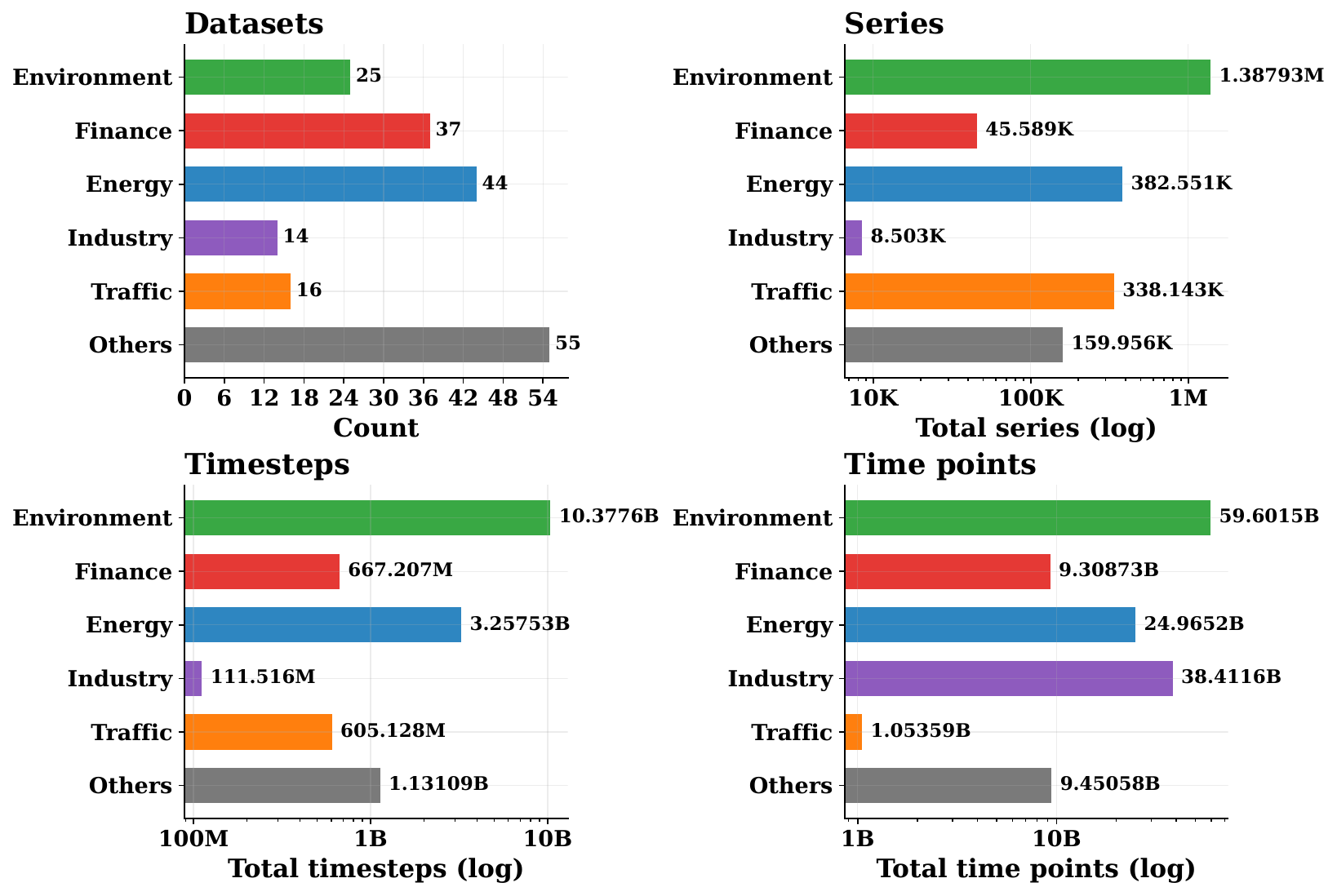}
    \caption{Domain-wise scale statistics of the proposed dataset.}
    \label{fig:statistics_1}
\end{figure}

\begin{figure}[htbp]
    \centering
    \includegraphics[width=0.65\linewidth]{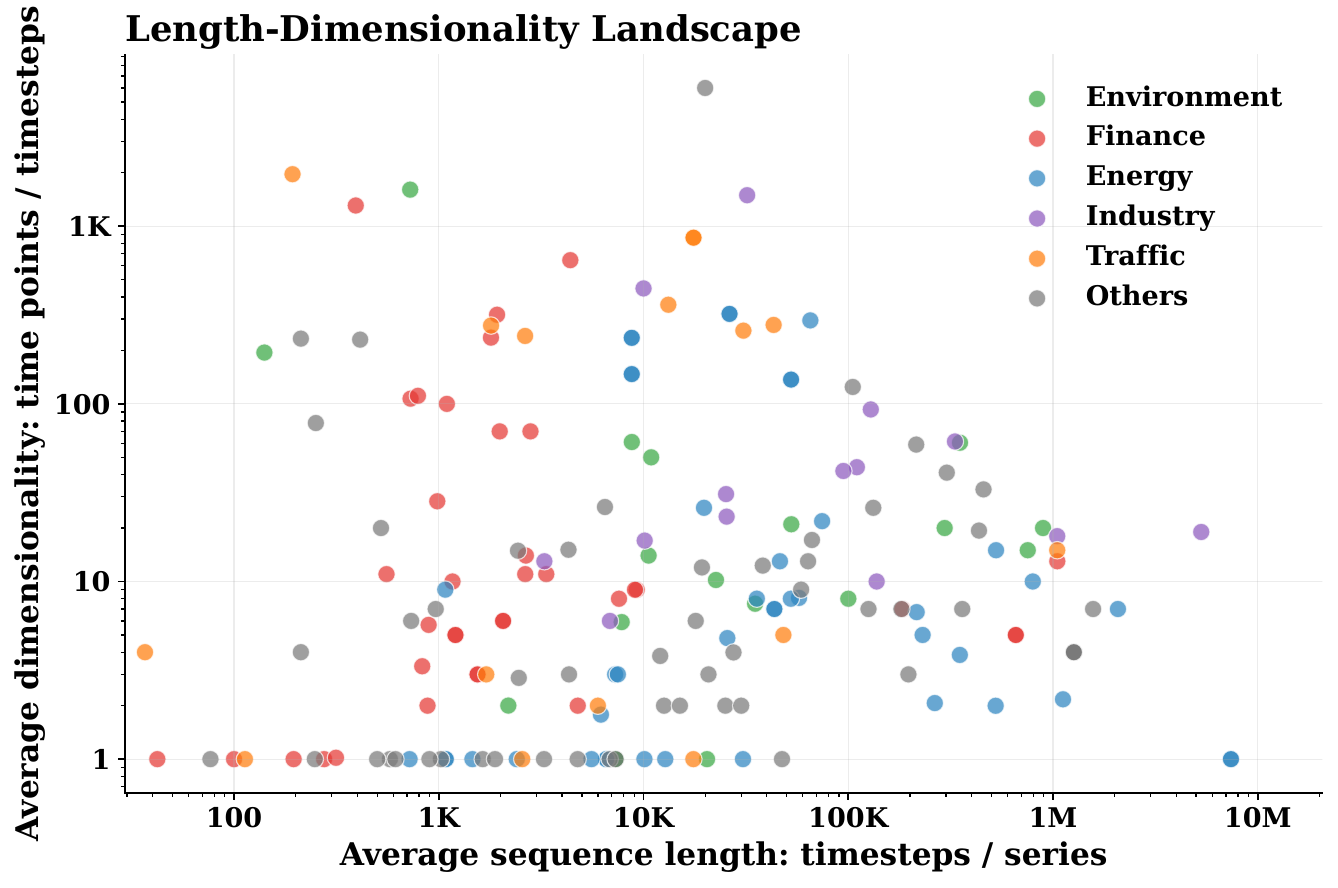}
    \caption{Length-dimensionality landscape of all subdatasets.}
    \label{fig:statistics_3}
\end{figure}

\begin{figure}[htbp]
    \centering
    \includegraphics[width=0.65\linewidth]{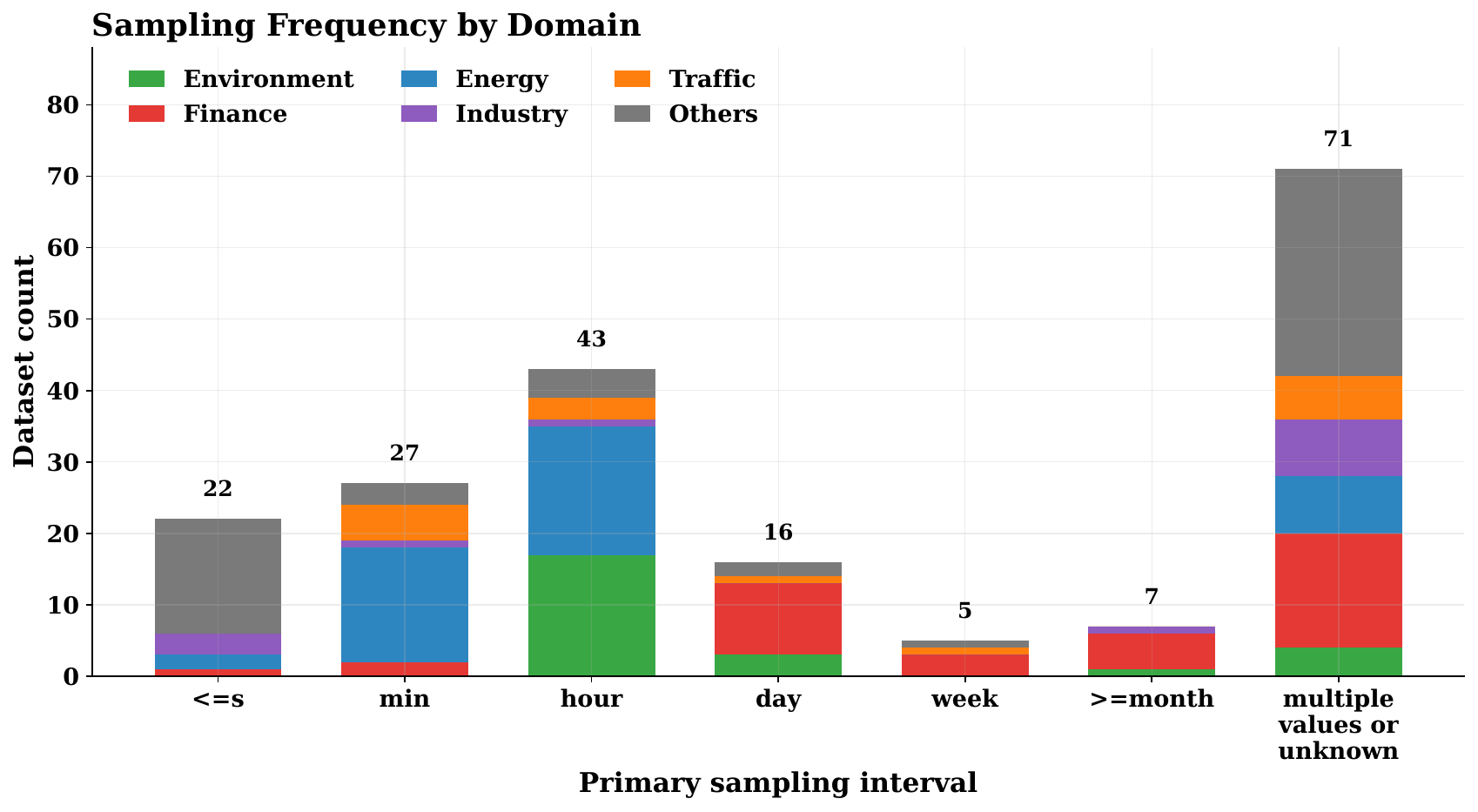}
    \caption{Sampling frequency distribution across domains.}
    \label{fig:statistics_4}
\end{figure}

\begin{figure}[htbp]
    \centering
    \includegraphics[width=0.65\linewidth]{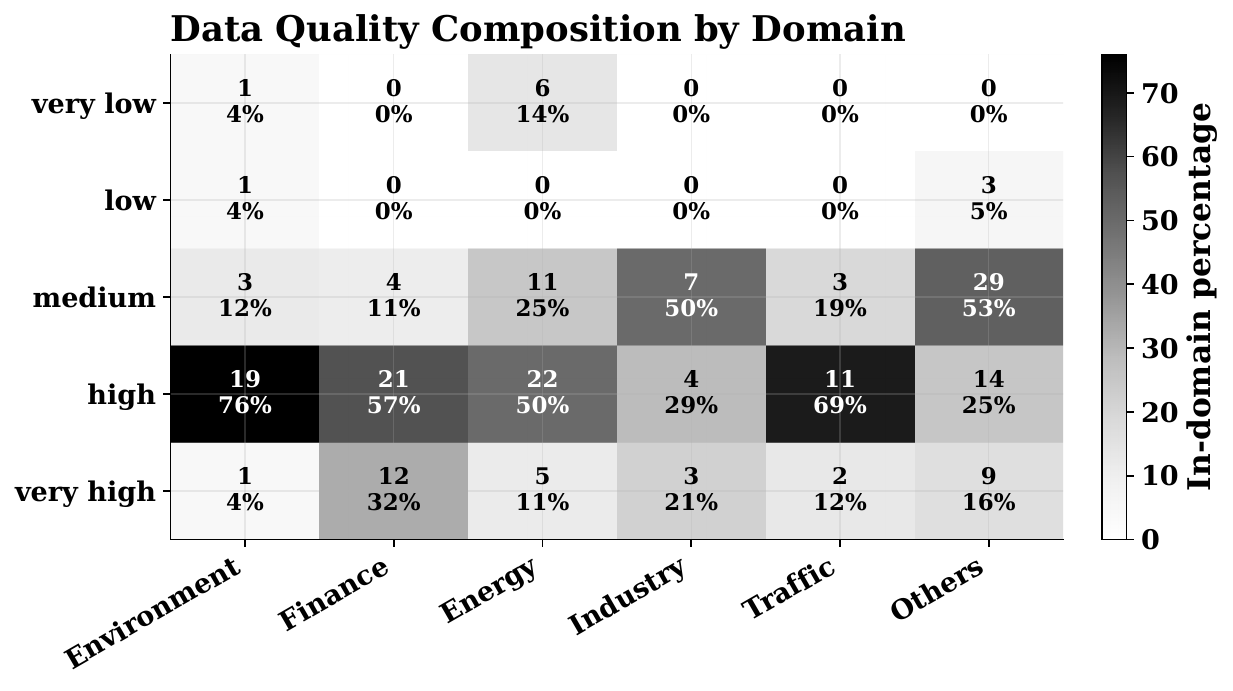}
    \caption{Data quality distribution across domains.}
    \label{fig:statistics_5}
\end{figure}

Figure~\ref{fig:statistics_1} provides an overview of the scale and domain distribution of the RMISC corpus, covering the number of subdatasets, time series, timesteps, and time points. These statistics demonstrate that RMISC offers large-scale real-world time series data while maintaining broad and relatively balanced coverage across diverse application domains. Figure~\ref{fig:statistics_3} presents the length-dimensionality landscape of all sub-datasets. The result shows that RMISC covers a wide range of sequence lengths and, more importantly, contains a substantial number of multivariate time series, including many high-dimensional datasets. Figure~\ref{fig:statistics_4} reports the sampling frequency distribution across different domains. The results show that RMISC covers a broad spectrum of temporal resolutions, ranging from sub-second and minute-level observations to hourly, daily, weekly, monthly, and lower-frequency records. This wide frequency coverage enables RMISC to support time series modeling under diverse temporal granularities. Figure~\ref{fig:statistics_5} summarizes the data quality distribution across domains. Data quality is assessed from both data-level and source-level perspectives. Specifically, we consider basic validity and usability indicators, such as duplicated or constant segments, abnormal values, and irregular records, as well as source-level factors, including the credibility of the original data platform and the clarity of metadata. The results show that the majority of subdatasets are of high or very high quality, indicating that RMISC provides a reliable foundation for large-scale time series pretraining and evaluation.

\section{Additional Experimental Results}\label{app:exp}

\subsection{Full Benchmark Results after Convergence}\label{subapp:full_benchmark}
Table~\ref{tab:benchmark_results} reports the detailed MASE and WQL results on four benchmarks.

\captionsetup[subtable]{labelformat=parens,labelsep=none,justification=centering}

\newcommand{\resultheader}{
\toprule
\multirow{2}{*}{\textbf{Training Corpus}}
& \multicolumn{2}{c}{\textbf{fev-uni}}
& \multicolumn{2}{c}{\textbf{fev-multi}}
& \multicolumn{2}{c}{\textbf{GIFT-uni}}
& \multicolumn{2}{c}{\textbf{GIFT-multi}} \\
\cmidrule(lr){2-3}
\cmidrule(lr){4-5}
\cmidrule(lr){6-7}
\cmidrule(lr){8-9}
& \textbf{MASE} & \textbf{WQL}
& \textbf{MASE} & \textbf{WQL}
& \textbf{MASE} & \textbf{WQL}
& \textbf{MASE} & \textbf{WQL} \\
\midrule
}
\begin{table*}[htbp]
\centering
\setlength{\tabcolsep}{1.5pt}
\renewcommand{\arraystretch}{0.90}

\caption{Out-of-distribution benchmark results of different training corpora on (a) Chronos-2, (b) GTT, (c) Moirai-2.0, and (d) TimesFM-2. Best results are highlighted in \textbf{bold}, and second best results are \underline{underlined}.}
\label{tab:benchmark_results}

\begin{subtable}{0.48\textwidth}
\centering
\caption{~Chronos-2}
\label{tab:chronos_results}
\resizebox{\linewidth}{!}{
\begin{tabular}{@{}lcccccccc@{}}
\resultheader
RU & 2.807 & 0.188 & 1.401 & 0.298 & 1.069 & 0.343 & 1.923 & 0.322 \\
SM & 2.214 & 0.180 & 1.039 & 0.227 & 1.003 & 0.320 & 1.597 & 0.307 \\
RM & 2.193 & 0.172 & 1.082 & 0.237 & 1.034 & 0.331 & 1.725 & \textbf{0.293} \\
RU+SM & 1.960 & 0.162 & 1.033 & \underline{0.217} & \textbf{0.937} & \textbf{0.308} & 1.670 & 0.306 \\
RU+RM & 2.290 & 0.167 & 1.070 & 0.234 & 1.000 & 0.321 & 1.716 & 0.302 \\
SM+RM & \underline{1.881} & \underline{0.159} & \textbf{0.972} & \textbf{0.205} & 1.007 & 0.327 & \underline{1.545} & 3.000 \\
RU+SM+RM & \textbf{1.775} & \textbf{0.154} & \underline{0.973} & \textbf{0.205} & \underline{0.973} & \underline{0.316} & \textbf{1.538} & \underline{0.299} \\
\bottomrule
\end{tabular}
}
\end{subtable}
\hfill
\begin{subtable}{0.48\textwidth}
\centering
\caption{~GTT}
\label{tab:gtt_results}
\resizebox{\linewidth}{!}{
\begin{tabular}{@{}lcccccccc@{}}
\resultheader
RU & 2.318 & 0.233 & 2.334 & 0.407 & 1.103 & \underline{0.372} & 2.032 & 0.423 \\
SM & 3.056 & 0.259 & 1.175 & 0.349 & 1.116 & 0.387 & 1.923 & 0.435 \\
RM & 2.598 & 0.244 & 1.165 & 0.322 & 1.094 & 0.380 & 1.953 & \underline{0.410} \\
RU+SM & 2.288 & \underline{0.232} & 1.143 & 0.339 & 1.093 & 0.377 & 1.916 & \textbf{0.409} \\
RU+RM & 2.358 & 0.234 & 1.150 & 0.329 & 1.099 & 0.374 & \underline{1.897} & 0.413 \\
SM+RM & \underline{2.283} & \textbf{0.227} & \underline{1.092} & \textbf{0.304} & \underline{1.039} & 0.421 & 1.900 & 0.458 \\
RU+SM+RM & \textbf{2.249} & \textbf{0.227} & \textbf{1.084} & \underline{0.315} & \textbf{1.038} & \textbf{0.362} & \textbf{1.791} & 0.415 \\
\bottomrule
\end{tabular}
}
\end{subtable}

\vspace{0.8em}

\begin{subtable}{0.48\textwidth}
\centering
\caption{~Moirai-2.0}
\label{tab:moirai_results}
\resizebox{\linewidth}{!}{
\begin{tabular}{@{}lcccccccc@{}}
\resultheader
RU & 2.156 & 0.178 & 1.095 & 0.256 & 1.057 & 0.344 & 1.792 & 0.334 \\
SM & 2.598 & 0.194 & 1.097 & 0.240 & 1.010 & 0.322 & 1.623 & 0.335 \\
RM & \underline{2.120} & \underline{0.175} & 1.034 & 0.225 & 1.040 & 0.332 & 1.601 & 0.313 \\
RU+SM & 2.250 & 0.177 & 1.041 & 0.223 & \textbf{0.972} & \underline{0.316} & 1.544 & \underline{0.308} \\
RU+RM & \textbf{2.026} & \textbf{0.169} & \textbf{0.994} & 0.221 & 0.988 & \textbf{0.315} & 1.639 & 0.321 \\
SM+RM & 2.308 & 0.178 & 1.044 & \textbf{0.212} & 1.005 & 0.321 & \underline{1.528} & \textbf{0.305} \\
RU+SM+RM & 2.142 & \underline{0.175} & \underline{1.016} & \underline{0.216} & \underline{0.979} & 0.321 & \textbf{1.515} & 0.324 \\
\bottomrule
\end{tabular}
}
\end{subtable}
\hfill
\begin{subtable}{0.48\textwidth}
\centering
\caption{~TimesFM-2.5}
\label{tab:timesfm_results}
\resizebox{\linewidth}{!}{
\begin{tabular}{@{}lcccccccc@{}}
\resultheader
RU & 2.300 & 0.203 & 1.156 & 0.284 & 1.109 & 0.360 & 2.071 & \underline{0.390} \\
SM & 2.337 & 0.195 & 1.084 & 0.256 & 1.087 & \underline{0.354} & 1.685 & \textbf{0.374} \\
RM & 2.117 & 0.192 & 1.061 & 0.257 & 1.134 & 0.368 & 1.732 & 0.392 \\
RU+SM & 2.161 & 0.192 & 1.100 & 0.255 & 1.117 & 0.357 & 1.749 & 0.392 \\
RU+RM & 2.074 & 0.190 & 1.046 & 0.254 & 1.098 & 0.361 & \underline{1.639} & 0.405 \\
SM+RM & \underline{2.056} & \underline{0.186} & \underline{1.038} & \underline{0.246} & \textbf{1.065} & \textbf{0.352} & 1.664 & 0.423 \\
RU+SM+RM & \textbf{1.980} & \textbf{0.183} & \textbf{1.033} & \textbf{0.241} & \underline{1.074} & 0.355 & \textbf{1.623} & 0.402 \\
\bottomrule
\end{tabular}
}
\end{subtable}
\end{table*}

\subsection{Benchmark MASE Results During Two-Epoch Pretraining}\label{subapp:benchmark_mase}

\begin{figure}
    \centering

    \begin{subfigure}{\linewidth}
        \centering
        \caption{Chronos-2}
        \label{subfig:chronos_benchmark_trend}
        \includegraphics[width=0.9\linewidth]{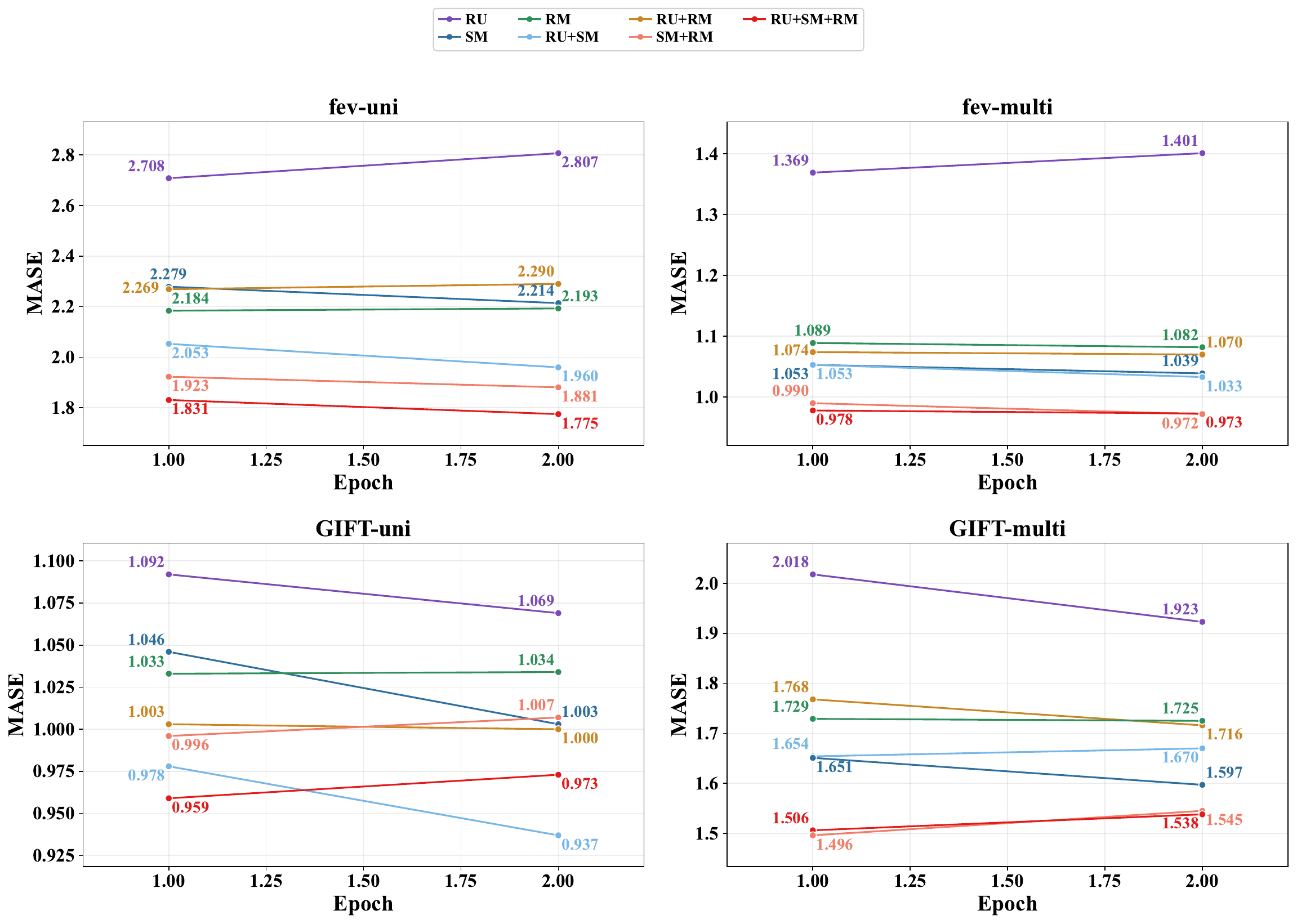}
    \end{subfigure}
    \hfill
    \begin{subfigure}{\linewidth}
        \centering
        \caption{GTT}
        \label{subfig:gtt_benchmark_trend}
        \includegraphics[width=0.9\linewidth]{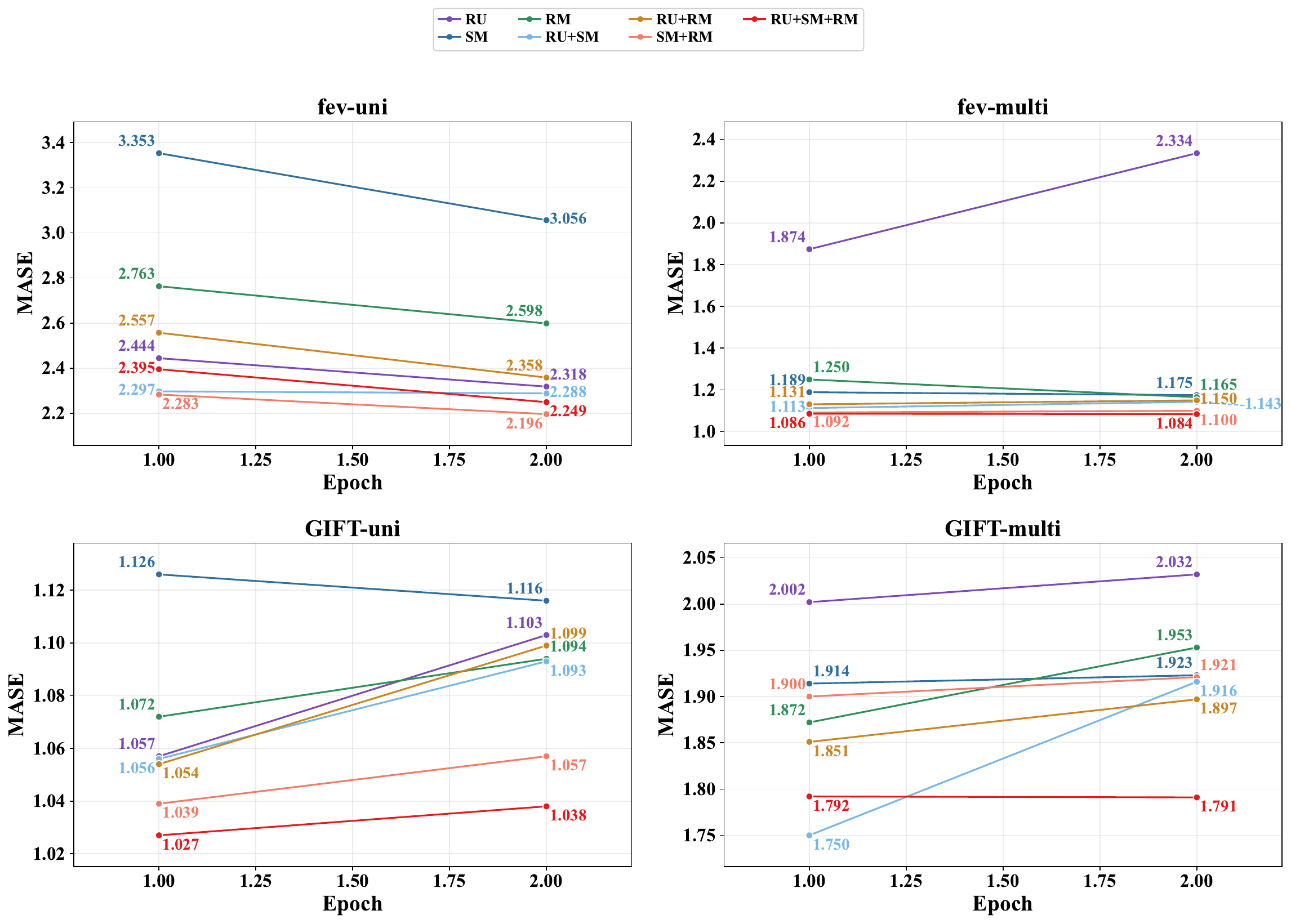}
    \end{subfigure}

    \caption{The changes in benchmark MASE scores from the first to the second epoch of different training corpora on (a) Chronos-2 and (b) GTT.}
    \label{fig:benchmark_trend}
\end{figure}

Figure~\ref{fig:benchmark_trend} reports changes in benchmark MASE scores from the first to the second epoch for Chronos-2 and GTT. For Chronos-2, although most corpora achieve lower MASE scores in the second epoch, the improvements are generally marginal. In addition, several corpora show increased MASE scores in the second epoch, suggesting a potential degradation in OOD generalization. For GTT, increases in MASE scores are more evident, as more corpora show higher MASE scores in the second epoch. Overall, these results suggest that both Chronos-2 and GTT converge by the end of the second epoch.

\end{document}